\documentclass[12pt]{article}    
\usepackage{fullpage}

\usepackage{amssymb}
\usepackage{graphicx}
\usepackage{amsmath}
\usepackage{url}
\usepackage{hyperref}
\usepackage{textcomp}
\usepackage{subfigure}
\usepackage{natbib}

\newtheorem{them}{Theorem}
\newtheorem{assum}{Assumption}

\newcommand{\argmin}{\mathop{\rm argmin}\limits}

\newcommand{\boldtheta}{{\boldsymbol{\theta}}}
\newcommand{\bolddelta}{{\boldsymbol{\delta}}}
\newcommand{\boldthetaP}{{\boldsymbol{\theta}}^{(p)}}
\newcommand{\boldthetaQ}{{\boldsymbol{\theta}}^{(q)}}
\newcommand{\boldthetaPtop}{{\boldsymbol{\theta}}^{(p)\top}}

\newcommand{\boldA}{{\boldsymbol{A}}}

\newcommand{\boldTheta}{{\boldsymbol{\Theta}}}

\newcommand{\boldu}{{\boldsymbol{u}}}

\newcommand{\boldSigma}{{\boldsymbol{\Sigma}}}

\newcommand{\boldDelta}{{\boldsymbol{\Delta}}}

\newcommand{\mathbbR}{\mathbb{R}}
\newcommand{\KL}{\mathrm{KL}}

\newcommand{\boldx}{{\boldsymbol{x}}}

\newcommand{\boldL}{{\boldsymbol{L}}}
\newcommand{\boldxp}{{\boldsymbol{x}}_{p}}
\newcommand{\boldxq}{{\boldsymbol{x}}_{q}}

\newcommand{\boldzero}{{\boldsymbol{0}}}

\newcommand{\boldThetaP}{{\boldsymbol{\Theta}^{(p)}}}
\newcommand{\boldThetaQ}{{\boldsymbol{\Theta}^{(q)}}}

\newcommand{\boldy}{{\boldsymbol{y}}}

\newcommand{\boldpsi}{{\boldsymbol{\psi}}}

\newcommand{\distP}{P}
\newcommand{\distQ}{Q}
\newcommand{\iid}{\stackrel{\mathrm{i.i.d.}}{\sim}}
\newcommand{\dx}{\mathrm{d}\boldx}

\newcommand{\vertiii}[1]{{\left\vert\kern-0.25ex\left\vert\kern-0.25ex\left\vert #1 
    \right\vert\kern-0.25ex\right\vert\kern-0.25ex\right\vert}}

\begin{document}

\title{Learning Sparse Structural Changes in High-dimensional Markov Networks: A Review on Methodologies and Theories}

\author{Song Liu \and Kenji Fukumizu \and Taiji Suzuki }

\author{Song Liu \\
              The Institute of Statistical Mathematics \\
              10-3 Midori-cho, Tachikawa, Tokyo 190-8562, Japan\\
              \url{liu@ism.ac.jp}           %
           \and
           Kenji Fukumizu \\
              The Institute of Statistical Mathematics\\
              10-3 Midori-cho, Tachikawa, Tokyo 190-8562, Japan\\
              \url{fukumizu@ism.ac.jp}
           \and 
           Taiji Suzuki \\
	          Tokyo Institute of Technology\\
	          2 Chome-12-1 Ookayama, Meguro, Tokyo 152-8550, Japan\\
	          PRESTO, Japan Science and Technological Agency (JST), Japan\\
	          \url{suzuki.t.ct@m.titech.ac.jp}
}

\maketitle

\begin{abstract}
	Recent years have seen an increasing popularity of learning the sparse \emph{changes} in Markov Networks. Changes in the structure of Markov Networks reflect alternations of interactions between random variables under different regimes and provide insights into the underlying system. While each individual network structure can be complicated and difficult to learn, the overall change from one network to another can be simple. This intuition gave birth to an approach that \emph{directly} learns the sparse changes without modelling and learning the individual (possibly dense) networks. In this paper, we review such a direct learning method with some latest developments along this line of research.
\end{abstract}

\section{Introduction}
\label{sec.intro}
The problem of learning the \emph{changes of interactions} between random variables can be useful in many applications. For example, genes may regulate each other in different ways when external conditions are changed; the number of daily flu-like symptom reports in nearby hospitals may become correlated when a major epidemic disease breaks out; EEG signals from different regions of the brain may be synchronized/desynchronized when the subject is performing different activities. Spotting such changes in interactions may provide key insights into the underlying system.

The interactions among random variables can be formulated as undirected probabilistic graphical models, or Markov Networks (MNs) \citep{Koller2009}, expressing the interactions via the \emph{conditional independence}. We  consider a simple model: the \emph{pairwise MNs} where the links are only encoded for single or pairs of random variables. Due to the Hammersley-Clifford theorem \citep{Hammersley1971}, the underlying joint probability density function can be represented as the product of univariate and bivariate factors.  

As an important challenge, structure learning of MNs has also attracted a significant amount of attention. Earlier methods \citep{Spirtes2000} use hypothesis testing to learn the conditional independence among random variables, which reflects the absences of edges. It is proved that such a problem is generally NP-hard \citep{Chickering1996}. Methods restricted to a sub-class of graphical models (such as trees or forests) \citep{Chow1968,Geman1984,Liu2011} also suffer from growing computational cost. 

However, the Hammersley-Clifford theorem together with the recent breakthrough on sparsity-inducing methods \citep{Tibshirani1996,Zhao2006,Wainwright2009} gave birth to many sparse structure learning ideas where the sparse factorization of the joint/conditional density function was estimated to infer the underlying structure of the  MN \citep{Friedman2008,Banerjee2008,Meinshausen2006,Ravikumar2010}. Although most works focused on parametric models, the structure learning has been conducted on semi-parametric ones in recent years. \citep{Liu2009,Liu2012}. 

There is also a trend of learning the \emph{changes} between MNs \citep{Zhang2010,Liu2014,Zhao2014}. Comparing to standard structure learning, the learning of changes views the problem in a more \emph{dynamic} fashion: Instead of estimating a static pattern, we hope to obtain a dynamic one, namely ``the change'' by comparing two sets of data. Since in some applications, the static pattern may not be computationally tractable, or simply too hard to comprehend. However, the difference between two patterns may be represented by some simple incremental effects involving only a small number of nodes or bonds, Thus it takes much less effort to learn and understand. 

One of the main uses of structural change learning is to spot responding variables in ``controlled experiments'' \citep{Zhang2010} where some key external factors of the experiments are altered, and two sets of samples are obtained. By discovering the changes in the MNs, we can see how random variables have responded to the change of the external stimuli. 

In this paper, we firstly review a recently proposed method of structural change learning between MNs \citep{Liu2014}. This follows a simple idea: if the MNs are products of the pairwise factors, the \emph{ratio} of two MNs must also be proportional to the ratios of those factors. Moreover, factors that do not change between two MNs will have no contribution to the ratio. This naturally suggests the idea of modelling the changes between two MNs $P$ and $Q$ 
as the ratio between two MN density functions $p(\boldx)$ and $q(\boldx)$. The ratio $p(\boldx)/q(\boldx)$ is directly estimated from a one-shot
estimation \citep{Sugiyama2012}. This density-ratio approach can work well
even when each MN is dense (as long as the change is sparse).

We also present some very recent theoretical results along this line of research. These works prove the \emph{consistency} of the density ratio method in the \emph{high-dimensional setting}. 
The \emph{support consistency} indicates the support of the estimated parameter converges to the support of the true parameter in probability. This is an important property for sparsity inducing methods.  
It is shown that under certain conditions the density ratio method recovers the correct parameter sparsity with high probability \citep{Liu2016a}. 
Moreover, \citeauthor{Fazayeli2016} introduced a theorem for the regularized density ratio estimator showing the estimation error, i.e., the $\ell_2$ distance between the estimated parameter and the true parameter converges to zero under milder conditions. 

As comparisons, we will also show a few alternative approaches to the change detection problem between MNs. The differential graphical model learning approach \citep{Zhao2014} uses a covariance-precision matrix equality to learn changes without going through the learning of the individual MNs. The ``jumping'' MNs \citep{Kolar2012} setting considers a scenario where the observations are received as a sequence and multiple sub-sequences are generated via different parametrizations of MN. 

We organize this paper as follows: Firstly, we introduce the problem formulation of learning changes between MNs in Section \ref{sec.dre}. Secondly, the density ratio approach and two other alternatives are explained in Section \ref{estimate.ratio.sec}. Section \ref{sec.learning.pg} reviews the theoretical results of these approaches. Synthetic and real-world experiments are conducted in Section \ref{sec.exp} to compare the performance of methods. Finally, in Section \ref{sec.open} and \ref{sec.concl}, we give a few possible future directions and conclude the current developments along this line of research. 

\section{Formulating Changes}
\label{sec.dre}
In this section, we focus on formulating the change of MNs using density ratio. At the end of this section, a few alternatives are also introduced. 
\subsection{Structural Changes by Parametric Differences}
\label{sec.prob.form}
Detecting changes naturally involves two sets of data. Consider independent samples drawn separately from two probability distributions $P$ and $Q$ on $\mathbb{R}^m$:
\begin{align*}
\mathcal{X}_p :=\{\boldx_p^{(i)}\}_{i=1}^{n_p} \iid P \text{ and }  \mathcal{X}_q :=\{\boldx_q^{(i)}\}_{i=1}^{n_q} \iid Q.
\end{align*}
We assume that $P$ and $Q$ belong to the family of \emph{Markov networks} (MNs)
consisting of univariate and bivariate factors,
i.e., 
their respective probability densities $p$ and $q$ are expressed as
\begin{align}
\label{eq.density.model}
p(\boldx;\boldthetaP) =\frac{1}{Z(\boldthetaP)}\exp\left(
\sum_{u,v = 1, u\ge v}^{m} \boldthetaP_{u,v}{}^\top \boldpsi_{u,v}(x_u,x_v) \right),
\end{align}
where
$\boldx = (x_{1}, \dots, x_{m})^\top$ is the $m$-dimensional random variable, $\top$ denotes the transpose,
$\boldthetaP_{u,v}$ is the $b$-dimensional parameter vector for the pair $(x_{u}, x_{v})$, and
\begin{align*}
\boldthetaP = (\boldthetaPtop_{1,1},\ldots, \boldthetaPtop_{m,1},\boldthetaPtop_{2,2},\ldots,\boldthetaPtop_{m,2},\ldots,\boldthetaPtop_{m,m})^\top
\end{align*}
is the entire parameter vector.
The feature function $\boldpsi_{u,v}(x_{u},x_{v})$ is a bivariate vector-valued basis function,
and $Z(\boldthetaP)$ is the normalization factor defined as
\begin{align*}
Z(\boldthetaP)  = \int \exp\left(\sum_{u,v = 1, u\ge v}^{m}  \boldthetaP_{u,v}{}^\top \boldpsi_{u,v}(x_{u},x_{v})\right)\dx.
\end{align*}
$q(\boldx; \boldthetaQ)$ is defined in the same way. Such a parametrization is generic when representing pairwise graphical models. 

Directly estimating an MN in this generic form is challenging since $Z(\boldthetaP)$ usually does not have a closed form except for a few special cases (e.g. Gaussian distribution). Markov Chain Monte Carlo \citep{Robert2005} is used to approximate such an integral. However, this would bring extra approximation errors. 

Nonetheless, we can define \emph{changes} between two MNs as the difference between their parameters. Therefore, given two parametric models $p(\boldx;\boldthetaP)$ and $q(\boldx;\boldthetaQ)$, we hope to discover \emph{changes in parameters} from $P$ to $Q$, i.e., \[\bolddelta = \boldthetaP-\boldthetaQ.\] Note that by its definition, the \emph{changes} are continuous. This is more advantageous than only considering  discrete changes of the MN structure, since a weak change of interaction does not necessarily shatter or flip the bond between two random variables. 

\subsection{Density Ratio Modelling}
An important observation is that
although two MNs may be complex individually, their changes might be ``simple'' since many terms may be cancelled while taking the difference, i.e. $\boldtheta^{(p)}_{u,v} -  \boldtheta^{(q)}_{u,v}$ might be zero. 
The key idea in \citep{Liu2014} is to consider the \emph{ratio} of $p$ and $q$:
\begin{align}
\label{eq.prop}
\frac{p(\boldx; \boldthetaP)}{q(\boldx; \boldthetaQ)}
\propto \exp \left( \sum_{u,v = 1, u\ge v}^m (\boldthetaP_{u,v}-\boldthetaQ_{u,v})^\top \boldpsi_{u,v}(x_{u},x_{v})\right),
\end{align}
where $\boldthetaP_{u,v}-\boldthetaQ_{u,v}$ encodes the difference
between $\distP$ and $\distQ$ for factor $\boldpsi_{u,v}(x_{u},x_{v})$, i.e., 
$\boldthetaP_{u,v} - \boldthetaQ_{u,v}$ is zero
if there is no change in the factor $\boldpsi_{u,v}(x_{u},x_{v})$.

Once the ratio of $p$ and $q$ is considered, 
each parameter $\boldthetaP_{u,v}$ and $\boldthetaQ_{u,v}$ does not have to be estimated.
Their difference $\bolddelta_{u,v}=\boldthetaP_{u,v} - \boldthetaQ_{u,v}$
is sufficient for change detection, as $\boldx$ only interacts with such a parametric difference in the ratio model.
Thus, in this density-ratio formulation,
$p$ and $q$ are no longer modelled separately,
but \emph{directly} as
\begin{align}
\label{ratio.model.def}
r(\boldx;\bolddelta) =
\frac{1}{N(\bolddelta)} \exp\left(\sum_{u,v = 1, u\ge v}^m \bolddelta_{u,v}^\top \boldpsi_{u,v}(x_{u},x_{v})\right),
\end{align}
where $N(\bolddelta)$ is the normalization term.
This direct formulation also halves the number of parameters
from both $\boldthetaP$ and $\boldthetaQ$ to only $\bolddelta$.

The normalization term $N(\bolddelta)$ is 
chosen to fulfill
$\int q(\boldx)r(\boldx;\bolddelta) \dx = 1$:
\begin{align*}
N(\bolddelta) = \int q(\boldx) \exp\left(\sum_{u,v = 1, u\ge v}^m \bolddelta_{u,v}^\top \boldpsi_{u,v}(x_{u},x_{v}) \right)\dx,  
\end{align*}
which is the expectation over $q(\boldx)$\footnote{If one models the ratio $\frac{q(x)}{p(x)}$, the normalization $$N(\bolddelta) = \int p(x) \exp\left(\sum_{u,v = 1, u\ge v}^m \bolddelta_{u,v}^\top \boldpsi_{u,v}(x_{u},x_{v}) \right) \dx$$  should be used.}. Note this integral is with respect to a true distribution where our samples are generated \footnote{$q(\boldx)$ should not be confused with $q(x;\boldtheta)$.}. 
This expectation form of the normalization term is another notable advantage
of the density-ratio formulation because it can be easily 
approximated by the sample average over $\{\boldxq^{(i)}\}_{i=1}^{n_q}\iid Q$:
\begin{align*}
&\hat{N}(\bolddelta; \boldx_q^{(1)}, \dots, \boldx_q^{(n_q)}) := \frac{1}{n_q}\sum_{i=1}^{n_q}
\exp\left(\sum_{u,v = 1, u\ge v}^m \bolddelta_{u,v}^\top \boldpsi_{u,v}(x_{q,u}^{(i)}, x_{q,v}^{(i)}) \right).
\end{align*}
Thus, one can always use this empirical normalization term 
for any (non-Gaussian) models $p(\boldx; \boldthetaP)$ and $q(\boldx; \boldthetaQ)$. 

Interestingly, if one uses $\psi_{u,v}(x_u x_v) = x_u x_v$ in the ratio model, it does not mean one assumes two individual MNs are Gaussian or Ising, it simply means we assume the changes of interactions are linear while other non-linear interactions remain unchanged. This formulation allows us to consider highly complicated MNs as long as their changes are ``simple''. 

Throughout the rest of the paper, we simplify the notation from $\boldpsi_{u,v}$ to $\boldpsi$ by assuming the feature functions are the same for all pairs of random variables.  

\subsection{Quasi Log-likelihood Equality}
\label{sec.quasi}
Density ratio is not the only direct modelling approach. Particularly for Gaussian MNs, where two distributions are parametrized as $p(\boldx;\boldTheta^{(p)})$ and $p(\boldx;\boldTheta^{(q)})$ with the precision matrix  $\boldTheta$, one alternative was proposed using the following equality \citep{Zhao2014}:
\begin{align}
\label{eq.quasi.log.likelihood}
\boldSigma^{(p)}\left(\boldTheta^{(p)} - \boldTheta^{(q)}\right)\boldSigma^{(q)} + \boldSigma^{(p)} - \boldSigma^{(q)} = \boldzero,
\end{align}
where $\boldSigma^{(p)}$ is the covariance matrix of the Gaussian distribution $p$. As we replace the covariance matrices $\boldSigma^{(p)}$ and $\boldSigma^{(q)}$ with their sample versions $\widehat{\boldSigma}^{(p)}$ and $\widehat{\boldSigma}^{(q)}$, it can be seen that $\boldTheta^{(p)} - \boldTheta^{(q)}$ is the only variable interacting with the data. Therefore, one may replace it with a single parameter $\boldDelta$ and later minimize the sample version of \eqref{eq.quasi.log.likelihood} (See Section \ref{sec.matching} for details). 

This direct formulation specifically uses a property of Gaussian MN that the covariance matrix computed from the data and the precision matrix that encodes the MN structure should approximately cancel each other when multiplied. However, such a relationship does not hold for other distributions in general. Studies on the generality of this equality is an interesting open question (See Section \ref{sec.open}). 

\paragraph{Remark} In fact, it is not necessary to combine $\boldthetaP - \boldthetaQ$ in \eqref{eq.prop} (or $\boldTheta^{(p)} - \boldTheta^{(q)}$ in \eqref{eq.quasi.log.likelihood}) into one parameter. However, such a model will be unidentifiable since there are too many combinations of $\boldTheta^{(p)}$ and $\boldTheta^{(q)}$ can produce the same difference. Nonetheless, such an indirect modelling may still be useful when the individual structures of the MNs are also our interests. We review an example of such indirect modelling in Section \ref{sec.joint.likelihood}.

\section{Learning Sparse Changes in Markov Networks} 
\label{estimate.ratio.sec}
\subsection{Density Ratio Estimation}
\emph{Density ratio estimation} has been recently introduced
to the machine learning community 
and is proven to be useful in a wide range of applications
\citep{Sugiyama2012}.
In \citep{Liu2014}, a density ratio estimator called 
the \emph{Kullback-Leibler importance estimation procedure} (KLIEP) 
for log-linear models \citep{Sugiyama2008a,Tsuboi2009} was employed
in learning structural changes.

For a density ratio model $r(\boldx; \bolddelta)$ (as introduced in \eqref{ratio.model.def}),
the KLIEP method minimizes the Kullback-Leibler divergence
from $p(\boldx)$ to $\hat{p}(\boldx;\bolddelta) = q(\boldx) r(\boldx;\bolddelta)$:
\begin{align}
\KL[p\|\hat{p}_\bolddelta] 
= \int p(\boldx) \log\frac{p(\boldx)}{q(\boldx)r(\boldx;\bolddelta)} \dx
=\text{Const.} - \int p(\boldx) \log r(\boldx; \bolddelta) \dx.
\label{eq.obj}
\end{align}
Note that the density ratio model \eqref{ratio.model.def} automatically
satisfies the non-negativity and normalization constraints:
\begin{align*}
r(\boldx;\bolddelta) > 0
~~\mbox{and}~~
\int q(\boldx) r(\boldx; \bolddelta) \dx = 1.
\end{align*}
Here we define \[\hat{r}(\boldx; \bolddelta) = \frac{
	\exp \left({\sum_{u,v = 1, u\ge v}^m \bolddelta_{u,v}^\top \boldpsi(x_{u},x_{v})}\right)}
{\hat{N}(\bolddelta; \boldx_q^{(1)}, \dots, \boldx_q^{(n_q)}) }\] as the \emph{empirical density ratio model}. In practice, one minimizes
the negative empirical approximation of the rightmost term in Eq.\eqref{eq.obj}:
\begin{align*}
\ell_{\mathrm{KLIEP}}(\bolddelta; \mathcal{X}_p, \mathcal{X}_q) &= 
-\frac{1}{n_p}\sum_{i=1}^{n_p} \log \hat{r}(\boldxp^{(i)}; \bolddelta)\\
&= - \frac{1}{n_p}\sum_{i=1}^{n_p} \sum_{u,v = 1, u\ge v}^m \bolddelta_{u,v}^\top \boldpsi(x_{p,u}^{(i)},x_{p,v}^{(i)}) \\
&+\log \left(\frac{1}{n_q}\sum_{i=1}^{n_q} \exp\left(\sum_{u,v = 1, u\ge v}^m \bolddelta_{u,v}^\top \boldpsi(x_{q,u}^{(i)},x_{q,v}^{(i)})\right)\right),
\end{align*}

\paragraph{Optimization} Since $\ell_{\mathrm{KLIEP}}(\bolddelta)$ consists of a linear part and a \emph{log-sum-exp} function \citep{Boyd2004}, it is convex with respect to $\bolddelta$, and
its global minimizer can be numerically found by standard optimization techniques
such as gradient descent.
The gradient of $\ell_{\mathrm{KLIEP}}$ with respect to $\bolddelta_{u,v}$ is given by
\begin{align}
\label{gradient.def}
\nabla_{\bolddelta_{u,v}} \ell_{\text{KLIEP}}(\bolddelta)
&=
-\frac{1}{n_p}\sum_{i=1}^{n_p} \boldpsi (x_{p,u}^{(i)},x_{p,v}^{(i)}) + \frac{1}{n_q} \sum_{i=1}^{n_q} \hat{r}(\boldx^{(i)}; \bolddelta) \boldpsi(x_{q,u}^{(i)},x_{q,v}^{(i)}),
\end{align}
that can be computed in a straightforward manner for \emph{any} feature vector $\boldpsi(x_{u},x_{v})$.

\subsection{Sparsity Inducing and Regularizations}
\label{sec.sparse.reg}
In the search for \emph{sparse} changes,
one may regularize the KLIEP solution
with a sparsity-inducing norm $\sum_{u\ge v} \| \bolddelta_{u,v} \|$, i.e.,
the \emph{group-lasso} penalty \citep{Yuan2006} where we use $\|\cdot\|$ to denote the $\ell_2$ norm.

Note that the density-ratio approach \citep{Liu2014}
directly sparsifies the difference $\boldtheta^{(p)}-\boldtheta^{(q)}$,
and thus intuitively this method can still work well even if $\boldtheta^{(p)}$ and $\boldtheta^{(q)}$ are dense
as long as $\boldtheta^{(p)}-\boldtheta^{(q)}$ is sparse. The following is the objective function used in \citep{Liu2014}:
\begin{align}
\label{eg.obj.final}
\hat{\bolddelta} = \argmin_{\bolddelta} \ell_{\text{KLIEP}}(\bolddelta) + \lambda \sum_{u,v = 1, u \ge v}^m\| \bolddelta_{u,v} \|.
\end{align}

In a recent work \citep{Fazayeli2016}, authors considered \emph{structured} changes, such as sparse, block sparse, node-perturbed sparse and so on. These structured changes can be represented via suitable atomic norms \citep{Chandrasekaran2012,Mohan2014}. For example, a KLIEP objective with a node-perturbation regularizer is
\begin{align}
\label{eg.obj.final.node}
\hat{\boldDelta} = \argmin_{\boldDelta\in \mathbb{R}^{m \times m}, L\in \mathbb{R}^{m\times m}} \ell_{\text{KLIEP}}(\boldDelta) + \lambda_1\| \boldDelta \|_1 + \lambda_2 \sum_{v=1}^m \left(\sum_{u=1}^m L_{u,v}^k\right)^{\frac{1}{k}}\notag\\
\text{subject to: } \boldDelta = \boldL + \boldL^\top,
\end{align}
Such a regularization can be used to discover perturbed nodes i.e., nodes that have a completely different connectivity pattern to other nodes among two networks.
\paragraph{Optimization}
Although the original objective of KLIEP was smooth and convex, the sparsity inducing norms are in general non-smooth. Proximal gradient methods, such as Fast Iterative Shrinkage Thresholding Algorithms (FISTA) \citep{Beck2009} can be utilized to solve regularized KLIEP objectives. A FISTA-like algorithm was proposed in \citep{Fazayeli2016} with a faster rate of convergence.

\subsection{Covariance-Precision Matching}
\label{sec.matching}
As mentioned above, the density ratio formulation is not the only way that may motivate the direct modelling. For the formulation using the equality \eqref{eq.quasi.log.likelihood}, we can solve the following sparsity inducing objective which was introduced in \citep{Zhao2014}.
\begin{align}
\label{eq.differential}
\hat{\boldDelta} = \argmin_{\boldDelta} \| \boldDelta \|_1 ~~ \text{subject to }
\| \hat{\boldSigma}^{(p)} \boldDelta \hat{\boldSigma}^{(q)} + \hat{\boldSigma}^{(p)} - \hat{\boldSigma}^{(q)}\|_{\infty} \le \epsilon,
\end{align}
where $\epsilon$ is a hyper-parameter. 
To obtain a sparse solution, we set a threshold for the solution at a certain level $\tau$, i.e. the value for $|\hat{\Delta}_{u,v}|<\tau$ is rounded to $0$. The constraint enforces the equality \eqref{eq.quasi.log.likelihood} and we used single parameter $\boldDelta$ replacing $\boldThetaP - \boldThetaQ$.

\paragraph{Optimization} This method is quite computationally demanding as the dimension $m$ grows. The Alternating Direction Method of Multipliers (ADMM) procedure \citep{Boyd2011} was implemented based on an augmented version of \eqref{eq.differential} (See Section 3.3 \citep{Zhao2014} for details).

\subsection{Maximizing Joint Likelihood}
\label{sec.joint.likelihood}
As it was mentioned in Section \ref{sec.quasi}, one does not have to use the direct modelling to learn sparse changes between MNs. In fact, separated modelling may not only discover changes, but also can recover the individual MN themselves. 
Recently,
a method based on \emph{fused-lasso}  \citep{Tibshirani2005}
has been developed \citep{Zhang2010}.
This method also sparsifies $\boldthetaP - \boldthetaQ$ directly.

The original method conducts \emph{feature-wise neighborhood regression} \citep{Meinshausen2006} jointly for $P$ and $Q$,
which can be conceptually understood as maximizing the local conditional Gaussian likelihood jointly on each random variable $t$.
A slightly more general form of the learning criterion may be summarized as
\begin{align}
\label{eq.fused}
\min_{\boldthetaP_{t} \in \mathbb{R}^{m-1}, \boldthetaQ_{t}\in \mathbb{R}^{m-1}} \ell_{t}(\boldthetaP_t; \mathcal{X}_p) &+ \ell_{t}(\boldthetaQ_t; \mathcal{X}_q)) \notag\\
&+ \lambda_1 (\|\boldthetaP_t\|_1+\|\boldthetaQ_t\|_1) + \lambda_2
\|\boldthetaP_{t}-\boldthetaQ_{t}\|_1,
\end{align}
where
$\ell_{t}(\boldtheta;\mathcal{X}_p)$ is the \emph{negative} log conditional likelihood
for the $t$-th random variable $x_t\in\mathbbR$ given the rest $\boldx_{\backslash t} \in\mathbbR^{m-1}$:
\begin{align*}
\ell_{t}(\boldtheta;\mathcal{X}_p) = -\frac{1}{n_p}\sum_{i=1}^{n_p}\log p(x_{p,t}^{(i)}|\boldx_{p,\backslash t}^{(i)};\boldtheta),
\end{align*} 
where each dimension of $\boldtheta$ corresponds to one of its potential neighborhood. $\ell_{t}(\boldtheta;\mathcal{X}_q)$ is defined in the same way as $\ell_{t}(\boldtheta;\mathcal{X}_p)$. 

Since the Fused-lasso-based method directly sparsifies the changes in MN structure,
it can work well even when each MN is not sparse (when $\lambda_1$ is set to 0).

\paragraph{Learning Changes in Sequence} Another recent development \citep{Kolar2012} along this line of research assumes the data points are received \emph{sequentially}, i.e., we observe 
$\boldx^{(1)}, \boldx^{(2)}, \dots, \boldx^{(T)}$ over time points $\mathcal{T} = \{1, 2, \dots, T\}$.
Suppose $\mathcal{T}$ can be segmented into $K$ disjoint unknown subsets: $\mathcal{T} = \cup_{k\in \left\{1\dots K\right\}}\mathcal{T}_k$ and $\boldx_{\mathcal{T}_k}  \sim p\left(\boldx, \boldtheta^{(\mathcal{T}_k)}\right)$.
The task is to segment such a sequence and learn an estimate $\widehat{\boldtheta}^{(\mathcal{T}_k)}$ for each segment. 
We can extend the idea of fused-lasso in \eqref{eq.fused}, and maximize the joint likelihood over each single observation:
\begin{align*}
\argmin_{\boldtheta^{(i)}, i \in \{1 \dots T\}} \sum_{i=1}^T \ell(\boldtheta^{(i)}; \boldx^{(i)}) + \lambda_1 \sum_{i=1}^T \left\|\boldtheta^{(i)}\right\|_1 + \lambda_2 \sum_{i=1}^{T-1} \left\| \boldtheta^{(i+1)} - \boldtheta^{(i)} \right\|_1,
\end{align*}
where the fused lasso term sparsifies the changes between MNs at adjacency time points, thus the learned $\boldtheta^{(1)}, \boldtheta^{(2)}, \dots \boldtheta^{(T)}$  is ``piecewise-constant'' and the segments are automatically determined from it. A block-coordinate descent procedure was proposed to solve this problem efficiently \citep{Kolar2010}. 

\section{Theoretical Analysis}
\label{sec.learning.pg}
The KLIEP algorithm does not only perform well in practice, it is also justified theoretically. In this section, we first introduce the support recovery theorem of KLIEP and then review some recent theoretical developments of direct change learning. 

\subsection{Preliminaries} In the previous section, a sub-vector of $\bolddelta$ indexed by a pair $(u,v)$ corresponds to a specific edge of an MN. From now on, we switch to a ``unitary'' index system as our analysis is not dependent on the edge nor the structure setting of the graph.

We introduce the ``true parameter'' notation $\bolddelta^*, p(\boldx)=q(\boldx)r(\boldx;\bolddelta^*),$
and the pairwise index set $E = \{(u,v) | u\ge v\}$.
Two sets of \textit{sub-vector indices} regarding to $\bolddelta^*$ and $E$ are  defined as  $S = \{t'\in E ~|~ \|\bolddelta^*_{t'}\| \neq 0\}, S^c = \{t'' \in E ~|~ \|\bolddelta^*_{t''}\| = 0\}.$ We rewrite the objective \eqref{eg.obj.final} as
\begin{align}
\label{eq.obj.alter}
\hat{\bolddelta} = \argmin_{\bolddelta} \ell_{\mathrm{KLIEP}}(\bolddelta) &+ \lambda_{n_p} \sum_{t\in S \cup S^c} \| \bolddelta_{t} \|.
\end{align}
Similarly we can define 
$\hat{S} = \{t' \in E ~|~ \|\hat{\bolddelta}_{t'}\| \neq 0\}$ and $\hat{S^c}$ accordingly.

Sample Fisher information matrix $\mathcal{I} \in \mathbb{R}^{\frac{b(m^2+m)}{2} \times \frac{b(m^2+m)}{2}}$ denotes the Hessian of the log-likelihood: $\mathcal{I} = \nabla^2 \ell_{\text{KLIEP}} (\bolddelta^*)$. $\mathcal{I}_{AB}$ is a sub-matrix of $\mathcal{I}$ indexed by two sets of indices $A, B \subseteq E$ are indices on rows and columns.

In this section, we prove the support consistency, i.e. with high probability that $S=\hat{S},S_c=\hat{S}_c$ (See e.g., Chapter 11 in \citep{Hastie2015} for an introduction of support consistency). 

\subsection{Assumptions}
We try \emph{not} to impose assumptions directly on each individual MNs, as the essence of KLIEP method is that it can handle various changes regardless the types of individual MNs.

The first two assumptions are essential to many support consistency theorems(e.g.  Eq. (15) and (16) in \citep{Wainwright2009}, Assumption A1 and A2 in \citep{Ravikumar2010}). These assumptions are made on the Fisher information matrix.

\begin{assum}[Dependency Assumption]
	\label{assum.depen}
	The sample Fisher information \textbf{submatrix}  $\mathcal{I}_{{SS}}$ has bounded eigenvalues:
	$
	\Lambda_\mathrm{min}(\mathcal{I}_{{SS}}) \ge \lambda_\mathrm{min} > 0,
	$
	with probability $1-\xi_q$, where $\Lambda_{\mathrm{min}}$ is the minimum-eigenvalue operator of a symmetric matrix.
\end{assum}
This assumption on the \emph{submatrix} of $\mathcal{I}$ is to ensure that the density ratio model is identifiable and the objective function is ``reasonably convex''.

\begin{assum}[Incoherence Assumption]
	\label{assum.incoherence}
	$
	\max_{t'' \in S^c}\|\mathcal{I}_{t''S} \mathcal{I}_{SS}^{-1}\|_1 \le 1-\alpha, 0<\alpha \le 1.
	$
	with probability 1, where $\|Y\|_1 = \sum_{i,j} \|Y_{i,j}\|_1$.
\end{assum}This assumption says the unchanged edges cannot exert overly strong effects on changed edges. Note this assumption is sometimes called ``\emph{irrepresentability}'' condition.

\begin{assum}[Smoothness Assumption on Likelihood Ratio]
	\label{assum.smooth}
	The log-likelihood ratio $\ell_\mathrm{KLIEP}(\bolddelta)$ is smooth around its optimal value, i.e., it has bounded derivatives
	\begin{align*}
	&\max_{\boldu, \|\boldu\|\leq \|\bolddelta^*\|}\left\| \nabla^2 \ell_\mathrm{KLIEP}(\bolddelta^*+\boldu)\right\|
	\leq \lambda_\mathrm{max} < \infty,\\
	&\max_{t\in S \cup S^c} \max_{\boldu, \|\boldu\|\leq \|\bolddelta^*\|}  \vertiii{\nabla_{\bolddelta_t}\nabla^2 \ell_\mathrm{KLIEP}(\bolddelta^* + \boldu)} 
	\leq \lambda_{3,\mathrm{max}}<\infty ,
	\end{align*}
	with probability $1$. 
\end{assum}
$\left\|\cdot\right\|$,  $\vertiii{\cdot}$ are the spectral norms of a matrix and a tensor respectively (See e.g., \citep{Tomioka2014} for the definition of spectral norm of a tensor).
This assumption guarantees the log-likelihood function is well-behaved. Now, we state the following assumptions on the density ratio:

\begin{assum}[Correct Model Assumption]
	\label{assum.correct}
	The density ratio model is correct, i.e. there exists $\bolddelta^*$ such that
	\begin{align*}
	p(\boldx) = r(\boldx;\bolddelta^*)q(\boldx).
	\end{align*}
\end{assum}
Although analyzing the mis-specified ratio model \citep{Kanamori2010} is certainly an interesting open question, we focus on correctly specified models in this section. 

\begin{assum}[Smooth Density Ratio Assumption]
	\label{assum.smooth.ratiomodel.nod}
	For any vector $\boldu \in \mathbb{R}^{\mathrm{dim}(\bolddelta^*)}$ such that $\|\boldu\|\leq \|\bolddelta^*\|$ and every $a\in \mathbb{R}$, the following inequality holds:
	\begin{align*}
	\mathbb{E}_q \left[\exp\left( a\left( r(\boldx, \bolddelta^* + \boldu) - 1 \right)\right) \right]  \le \exp\left(Ma^2\right),
	\end{align*}
\end{assum}
where $M>0$ is a constant independent from $m$. 
This assumption states that the density ratio model, around its optimal parameter, should not often obtain large values over samples from $Q$.

\subsection{Successful Support Recovery of KLIEP \citep{Liu2016a,Liu2016}}
\label{sec.them.kliep}
\begin{them}
	\label{them.the.main.theorem}
	Suppose that Assumptions \ref{assum.depen}, \ref{assum.incoherence}, \ref{assum.smooth}, \ref{assum.correct}, and \ref{assum.smooth.ratiomodel.nod} as well as
	\begin{align}
	\label{eq.true.para.condition}
	\min_{t'\in S} \|\bolddelta^*_{t'}\| \geq \frac{10}{\lambda_\mathrm{min}} \sqrt{d}\lambda_{n_p}	
	\end{align}
	are satisfied, where $d$ is the number of changed edges defined as $d = |S|$, i.e., the cardinality of the set of non-zero parameter groups.
	Suppose also that the regularization parameter is chosen so that
	\begin{align}
	\label{eq.lambda.condition}
	M_1 \sqrt\frac{{\log \frac{m^2+m}{2}}}{n_p} \le \lambda_{n_p} \le M_2 \min\left(\frac{\|\bolddelta^*\|}{\sqrt{b}}, 1\right),
	\end{align}
	and $n_q \ge M_3 n_p^2$, where $M_1, M_2$ and $M_3$ are constants. 
	Then there exist some constants $L_1$,  $K_1$, and $K_2$ such that if $n_p\geq L_1 d^2\log  \frac{m^2+m}{2}$, with the probability at least
	\begin{align*}
	1- \exp\left( - K_1 \lambda_{n_p}^2n_p \right) - 4\exp\left( -K_2 dn_q \lambda_{n_p}^4 \right) - \xi_q,
	\end{align*}
	the following properties hold:
	\begin{itemize}
		\item Unique Solution: The solution of \eqref{eq.obj.alter} is unique.
		\item Successful Change Detection: $\hat{S} = S$ and $\hat{S}^c = S^c$.
	\end{itemize}
\end{them}

The proof of this theorem follows the Primal-dual witness construction (See e.g., Section 11.4.2 in \citet{Hastie2015}). 

\paragraph{Remark} The main conclusion of this theorem states that if the regularization parameter is \emph{reasonably chosen} \eqref{eq.lambda.condition} and the true non-zero groups $\|\bolddelta^*_{t'}\|, {t'}\in S$ is \emph{large enough} \eqref{eq.true.para.condition}, with high probability, we are guaranteed to have the correct support of parameters. The samples needed for $n_p$ only grows linearly with $\log m$ and $n_q$ grows quadratically with $n_p$.
\subsection{$\ell_2$ Consistency of KLIEP with Atomic Norm \citep{Fazayeli2016}}
\label{sec.consis.atomic}
As it was introduced in Section \ref{sec.sparse.reg}, atomic norms can be used to learn changes with special topological structures. Instead of support recovery, we focus on the $\ell_2$ loss between the estimated parameter $\hat{\bolddelta}$ and the true parameter $\bolddelta^*$, i.e., $\|\bolddelta^* - \hat{\bolddelta}\|$. 

First, we generalize our objective function as 
\begin{align}
\label{eq.obj.alter.2}
\hat{\bolddelta} = \argmin_{\bolddelta \in \mathbb{R}^{\frac{m^2+m}{2}}} \ell_{\mathrm{KLIEP}}(\bolddelta) &+  \lambda_{n_p,n_q} R(\bolddelta),
\end{align}
where $R$ is an atomic norm function.

Such a theorem relies on the \emph{Restricted Strong Convex (RSC)} property on the \emph{Error Set} of the objective function. Intuitively, if $\ell_{\mathrm{KLIEP}}(\bolddelta)$ is ``highly curved'', small $|\ell_{\mathrm{KLIEP}}(\hat{\bolddelta}) - \ell_{\mathrm{KLIEP}}(\bolddelta^*)|$  ensures small $\|\hat{\bolddelta} - \bolddelta^*\|$. Thus we only need to figure out how $|\ell_{\mathrm{KLIEP}}(\hat{\bolddelta}) - \ell_{\mathrm{KLIEP}}(\bolddelta^*)|$ reaches zero as number of samples goes to infinity and this is a more accessible target. 

To make sure our objective has such a ``strongly convex'' curvature, one need to impose a uniform lower-bound on the eigenvalues of the objective Hessian (a.k.a., sample Fisher information matrix $\mathcal{I}$). However, this is not realistic for the high-dimensional setting, as $\mathcal{I}$ is certainly rank-deficient. As an alternative, we impose an assumption on the convexity of $\ell_\text{KLIEP}$ over a constrained set:
\paragraph{Restricted Strong Convex Condition} The function $\ell$ is Restricted Strong Convex (RSC) at a cone $C$ if there exists a constant $\kappa$ such that $\forall \boldu \in C$
\begin{align*}
\ell(\bolddelta^* + \boldu) - \ell(\bolddelta^*) - \langle 
\boldu, \nabla_\ell(\bolddelta^*)\rangle \ge \kappa \|\boldu\|^2.
\end{align*}

If $\bolddelta^* - \hat{\bolddelta} \in C$, it is possible to obtain a \emph{deterministic} bound (Theorem 2 in \citet{Banerjee2014}) on the $\ell_2$ estimation error
\begin{align*}
\|\bolddelta^* - \hat{\bolddelta}\|_2 = O\left(\frac{\lambda_{n_p, n_q}}{\kappa}\Psi(C)\right),
\end{align*}
where $\Psi(C)$ is the the norm compatibility constant \citep{Negahban2009} and can be easily bounded.
Note that although this bound itself is not probabilistic, the parameter $\lambda_{n_p, n_q}$ is random and the RSC may hold with a probability. One can infer the sample complexity from these bounds.

Two things remain to be shown. First, we need to find such a cone which contains $\hat{\bolddelta} - \bolddelta^*$. Second, we need to prove $\ell_\text{KLIEP}$ is RSC on this cone. We start with the first problem.

\paragraph{Error Set (Lemma 1 in \citep{Banerjee2014})} For any convex loss $\ell(\bolddelta)$, if $\lambda_{n_p, n_q}$ is large enough, i.e.,  \[\lambda_{n_p, n_q}\ge \beta R^*(\nabla \ell(\bolddelta^*) ), \beta > 1 \] where $R^*$ is the dual norm of $R$, 
it can be proven that the estimation error $\boldu = \bolddelta^* - \hat{\bolddelta}$ lies in an Error Set:
\begin{align*}
E_r = \left\{ \boldu, \boldu \in \mathrm{dom}(\bolddelta) \bigg| R(\bolddelta^* + \boldu)
\le R(\bolddelta^*) + \frac{1}{\beta} R(\boldu) \right\},
\end{align*} 
where $\mathrm{dom}(\boldy)$ is the domain of $\boldy$. Let's define $C_r = \mathrm{cone}(E_r)$.

In fact, it can be shown that if \[\lambda_{n_p. n_q} \ge \frac{c\cdot (w(\Omega_R) + \epsilon)}{\sqrt{\min(n_p, n_q)}},\] where $w(A)$ is the Gaussian width of a set $A$ \citep{Ledoux2013} and $\Omega_R = \{\boldu | R(\boldu) \le 1\}$, then $\lambda_{n_p, n_q}\ge \beta R^*(\nabla \ell(\bolddelta^*) )$ holds automatically with high probability (Theorem 1 in \citep{Fazayeli2016}). Now we have a cone $C_r$ where $\hat{\bolddelta} - \bolddelta^*$ resides.

As to the second problem, it can be proven that $\ell_{\mathrm{KLIEP}}$ is RSC at $C_r$ with high probability once $n_q \ge n_0, n_0 = w^2(C_r \cap S)$, where $S$ is a unit hypersphere (Theorem 2 in \citep{Fazayeli2016}). Thus $n_0$ is the minimum number of samples required from $Q$ to be able to apply this theorem.  

Putting everything together, we have the main theorem proved in \citep{Fazayeli2016}:
\begin{them}[$\ell_2$ Consistency of Atomic Norms]
	If Assumption \ref{assum.smooth.ratiomodel.nod} holds, and $\hat{\bolddelta}$ is the minimizer of \eqref{eq.obj.alter.2}, then with probability at least $1 - M_1\mathrm{exp}(-\epsilon^2)$ the followings hold: 
	\begin{align*}
	\lambda_{n_p, n_q} \ge \frac{M_2}{\sqrt{\min(n_p, n_q)}} (w(\Omega_R)+\epsilon))
	\end{align*}	
	and for $n_q \ge c_1 w^2 (C_r \cap S)$, with high probability, the estimate $\hat{\bolddelta}$ satisfies 
	\begin{align*}
	\|\hat{\bolddelta} - \bolddelta^* \|_2 = O\left(\frac{w(\Omega_R)}{\sqrt{\min(n_p, n_q)}}\right)\Psi(C_r)
	\end{align*}
\end{them}
Note the constants $M_1$ and $M_2$ listed in this theorem are not the same as the ones in Theorem \ref{them.the.main.theorem}. 
To apply this theorem, we need to know the bounds of $w(\Omega_R)$ and $\Psi(C_r)$ for specific $R$ norms. These bounds have been proven in previous literatures (see e.g. \citep{Banerjee2014}). For example, if $R$ is $\ell_1$ norm, then $\Psi(C_r) \le 4\sqrt{d}$ and $w(\Omega_R) \le c \log m$ so applying the above theorem, we have 
\[\|\hat{\bolddelta} - \bolddelta^* \|_2 = O\left(\sqrt{\frac{d\log m}{\min(n_p, n_q)}}\right). \] 

\paragraph{Remark} Although this bound does not directly prove the support consistency, we can learn that sample complexity $\min(n_p, n_q) = \Omega( d\log m)$ guarantees the convergence of estimation error in $\ell_2$ norm. As to $n_q$, it should also satisfy $n_q \ge c_1 w^2(C_r\cap S)$, which is again $n_q = \Omega( d \log m )$ in the case of $\ell_1$ norm. This sample complexity is milder than what \citeauthor{Liu2016a} have obtained in the previous section $\Omega(d^2 \log (m^2 + m)/2)$ and $n_q = \Omega(n_p^2)$. Nonetheless, both theories can be applied to high dimensional regime $m\gg \min(n_p, n_q)$.

\subsection{Support Consistency of Covariance-Precision Matching \citep{Zhao2014}} 
\label{sec.convar.prec.theorem}
In this section, we introduce the support recovery theorem of the Covariance-Precision Matching method \eqref{eq.differential} in terms of support consistency on Gaussian MNs. Specifically for Gaussian MNs, we need a slightly different set of notations, as they are parametrized in matrix forms. $\Sigma_{j,k}^{(p)}$ is the $j,k$-th elements of matrix $\boldSigma^{(p)}_{j,k}$ and $\Sigma_\mathrm{max}^{(p)}$ is $\max_j \Sigma^{(p)}_{j,j}$.

The first assumption is to ensure that the ``amount of change'' is fixed and the change is always sparse, and does not grow with the number of dimension $m$.

\begin{assum}
	\label{assum.diff.1}
	The difference matrix $\boldDelta$ has $d \le m$ non-zero elements in its upper triangular sub-matrix. $|\boldDelta|_1 \le M_0$, and both $d$ and $M_0$ does not depend on dimension $m$.
\end{assum}

The second assumption assures that the covariates are not strongly dependent if there are many changes in the precision matrix. This is similar to the incoherence assumption used in Assumption \ref{assum.incoherence}. 
\begin{assum}
	\label{assum.diff.2}
	The constants $\mu^{(p)} = \max_{j\neq k} |\Sigma_{j,k}^{(p)}|$ and $\mu^{(q)} = \max_{j\neq k} |\Sigma_{j,k}^{(q)}|$ must satisfy $\mu = 4\max(\mu^{(p)}\Sigma_{\mathrm{max}}^{(q)},
	\mu^{(q)}\Sigma_{\mathrm{max}}^{(p)}) \le \frac{\Sigma^S_{\mathrm{min}}}{2d} $, where \[\Sigma^S_{\mathrm{min}} = \mathrm{min}_{j,k}(\Sigma^{(q)}_{jj}\Sigma^{(p)}_{jj}, \Sigma^{(q)}_{kk}\Sigma^{(p)}_{jj} + 2\Sigma^{(q)}_{kj}\Sigma^{(p)}_{jk} + \Sigma^{(q)}_{jj}\Sigma^{(p)}_{kk}).\]
\end{assum}

We first intuitively explain how the proof works. The proof of the support consistency can be thought as controlling $\|\hat{\boldDelta} - \boldDelta^*\|_\infty$.  Clearly, for the population covariance matrices $\boldSigma^{(p)}$ and $\boldSigma^{(q)},$
$\boldSigma^{(p)}\boldDelta^*\boldSigma^{(q)} + \boldSigma^{(p)} - \boldSigma^{(q)} = \boldzero$. If we replace the above population covariances with their sample versions, we can expect $\|\hat{\boldSigma}^{(p)}\boldDelta^*\hat{\boldSigma}^{(q)} + \hat{\boldSigma}^{(p)} - \hat{\boldSigma}^{(q)}\|_\infty \le \epsilon,$ if number of samples is large enough. Furthermore, $\epsilon$ can be a function decreasing with $\min(n_p, n_q)$ as the estimated covariances are getting closer and closer to the population ones. 

Therefore, if we set the $\epsilon$ to a decreasing function, we can still ``contain'' the optimal parameter $\boldDelta^*$ in the feasible zone with high probability. By definition, the estimated difference $\hat{\boldDelta}$ should also be in the feasible zone, thus they should not be far off, if the zone is small enough. The rigorous proof of the above statements is given in the Appendix of \citet{Zhao2014}.

Now, we give the support recovery theorem\footnote{In fact, the support recovery theorem was proved for a slightly augmented version of \eqref{eq.differential}.}  as follows (See Section 4 in \citep{Zhao2014} for details): 
\begin{them}[Support Consistency of Covariance-Precision Matching]
	Suppose $P$ and $Q$ are Gaussian, Assumption \ref{assum.diff.1} and \ref{assum.diff.2} hold,   $\mathrm{min}(n_p, n_q) \ge \log m$ and 
	\[\tau_{n_p, n_q} = \Omega \left(\sqrt{\frac{\log m}{\min(n_p, n_q)}}\right), \epsilon_{n_p, n_q} =  M_1 \cdot \sqrt{\frac{\log m}{\min(n_p, n_q)}}\]
	and $\min_{j,k} |\Delta^*_{j,k: \Delta^*_{j,k} \neq 0}| \ge 2\tau_{n_p, n_q}$\footnote{$\tau_{n_p, n_q}, \epsilon_{n_p,n_q}$ is the sample-dependent version of $\tau,\epsilon$ introduced in Section \ref{sec.matching}.}, then with high probability, \eqref{eq.differential} can recover the correct support of $\boldDelta^*$.
\end{them}
This support consistency theorem, although only applies to Gaussian MNs, has similar structure to the one derived for KLIEP (Section \ref{sec.them.kliep}). First, they both assume the true non-zero parameter should be large enough. Second, they both assume the sparsity inducing factor ($\lambda_{n_p, n_q}$ and $\tau_{n_p, n_q}$) should decay as the sample size $\min(n_p, n_q)$ increases, while increase as the log-dimension $\log m$ increases.

\subsection{Summary and Discussion}
\label{sec.summary}
Now, we summarize and compare these theoretical results. First we discuss the similarities of these theorems. 
\begin{itemize}
	\item None of the above proofs require the sparsity assumption on each individual MN. Thus in theory, all methods should work well when individual MNs are dense. 
	\item The efficiency of all methods are affected by the sparsity of changes (i.e. $d$). This make sense since the sparsity assumption is made on the changes between two MNs. 
	\item All theorems apply to the high dimensional regime ($m \gg \min(n_p, n_q)$). None requires $n_p$ \emph{or} $n_q$ to be comparable to the dimensionality $m$.
\end{itemize}
However, there is one important difference among these theorems. The sample complexities introduced in Section \ref{sec.them.kliep} and \ref{sec.consis.atomic} are \emph{not} symmetric. the sample complexity of $n_q$ is more restrictive comparing to that of $n_p$. This is understandable since KLIEP itself is an asymmetric method (KL divergence is asymmetric). In comparison, the sample complexity of Covariance-Precision Matching is symmetric, i.e., the theorem does not show the ``bias'' toward either of the datasets. Thus, if one has perfectly balanced Gaussian datasets, it might be natural to use Covariance-Precision Matching to learn the differences. 

\section{Experiments}
\label{sec.exp}
In this section, we compare the performance of two direct change detection methods: KLIEP and Covariance-Precision (CP) Matching using synthetic and real-world examples.

\subsection{Implementations}
Sparsity-inducing KLIEP can be implemented using sub-gradient descent approach. The MATLAB\textregistered code can be found at \url{http://www.ism.ac.jp/~liu/kliep_sparse/demo_sparse.html}.

The R \citep{RCoreTeam2016} implementation of CP matching using ADMM can be obtained at \url{https://github.com/sdzhao/dpm}.

\subsection{Synthetic Examples}
\begin{figure*}
	\centering
	\subfigure[The ground truth. $m=50, d = 6$.]{
		\label{fig.toy.ground}
		\includegraphics[width=.5\textwidth]{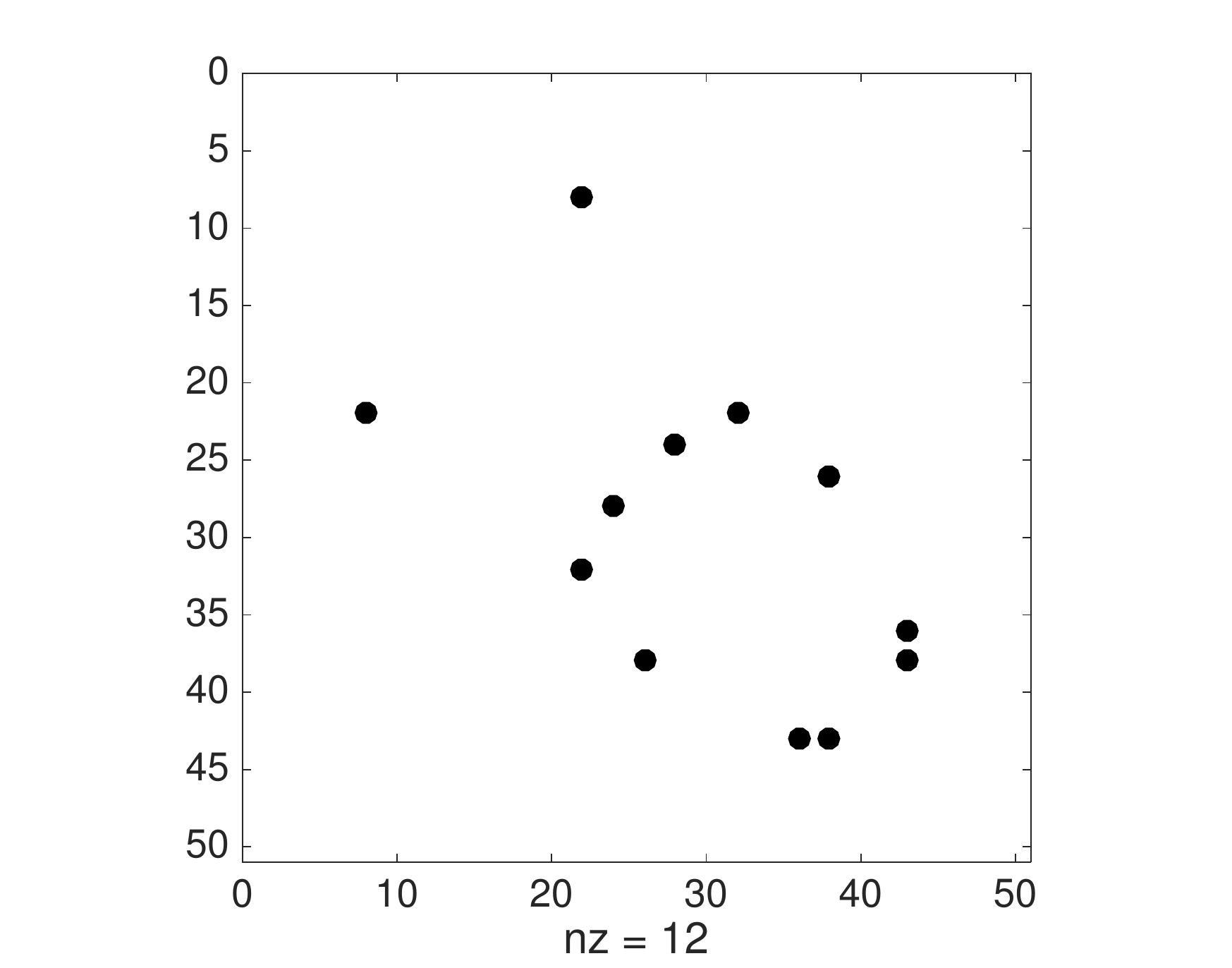}}
	\subfigure[ROC curves]{
		\label{fig.toy.roc}
		\includegraphics[width=.45\textwidth]{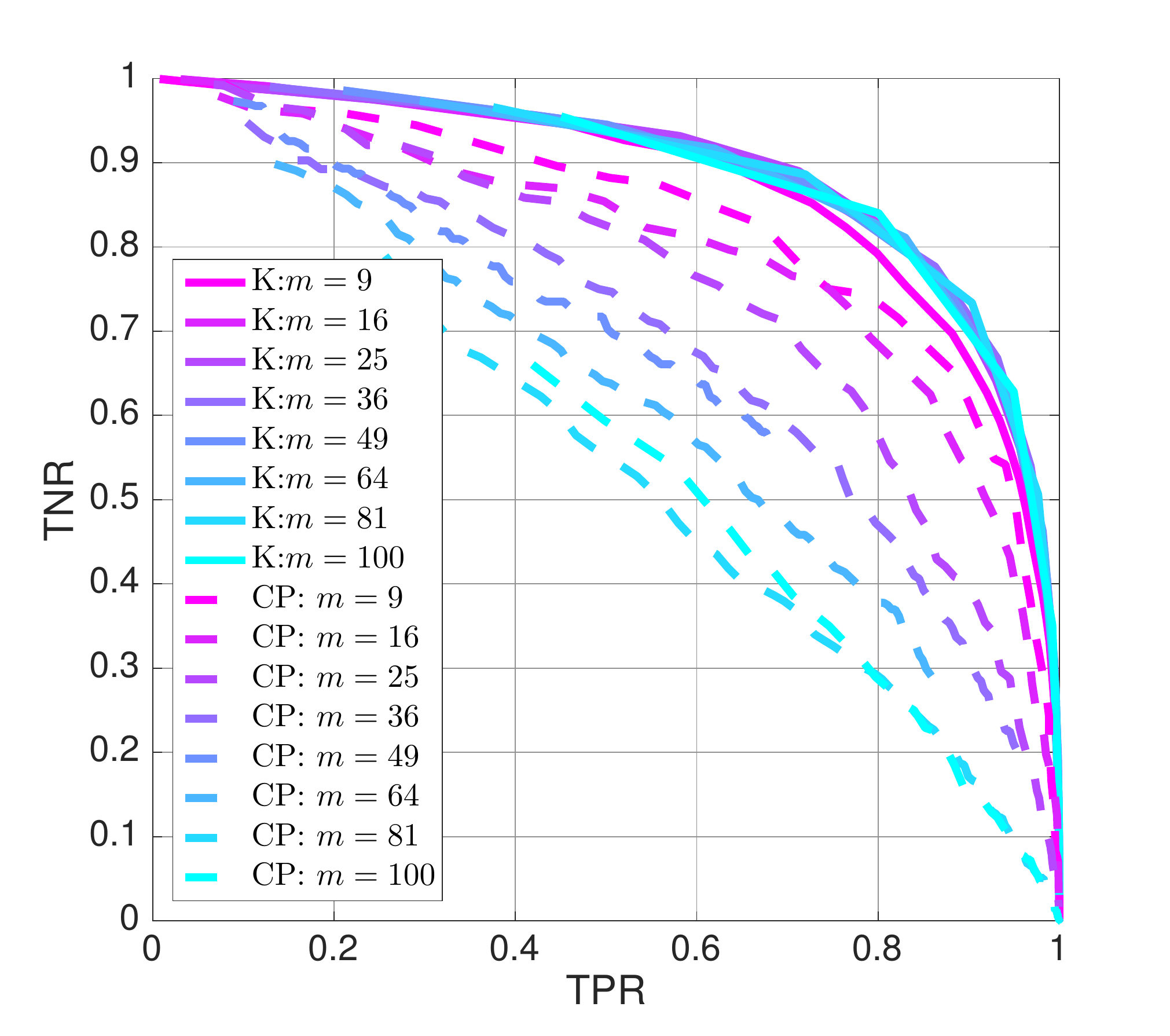}
	}
	\subfigure[$\hat{\boldDelta}$, KLIEP, $\alpha = .75$ ]{
		\label{fig.toy.kliep.75}
		\includegraphics[width=.3\textwidth]{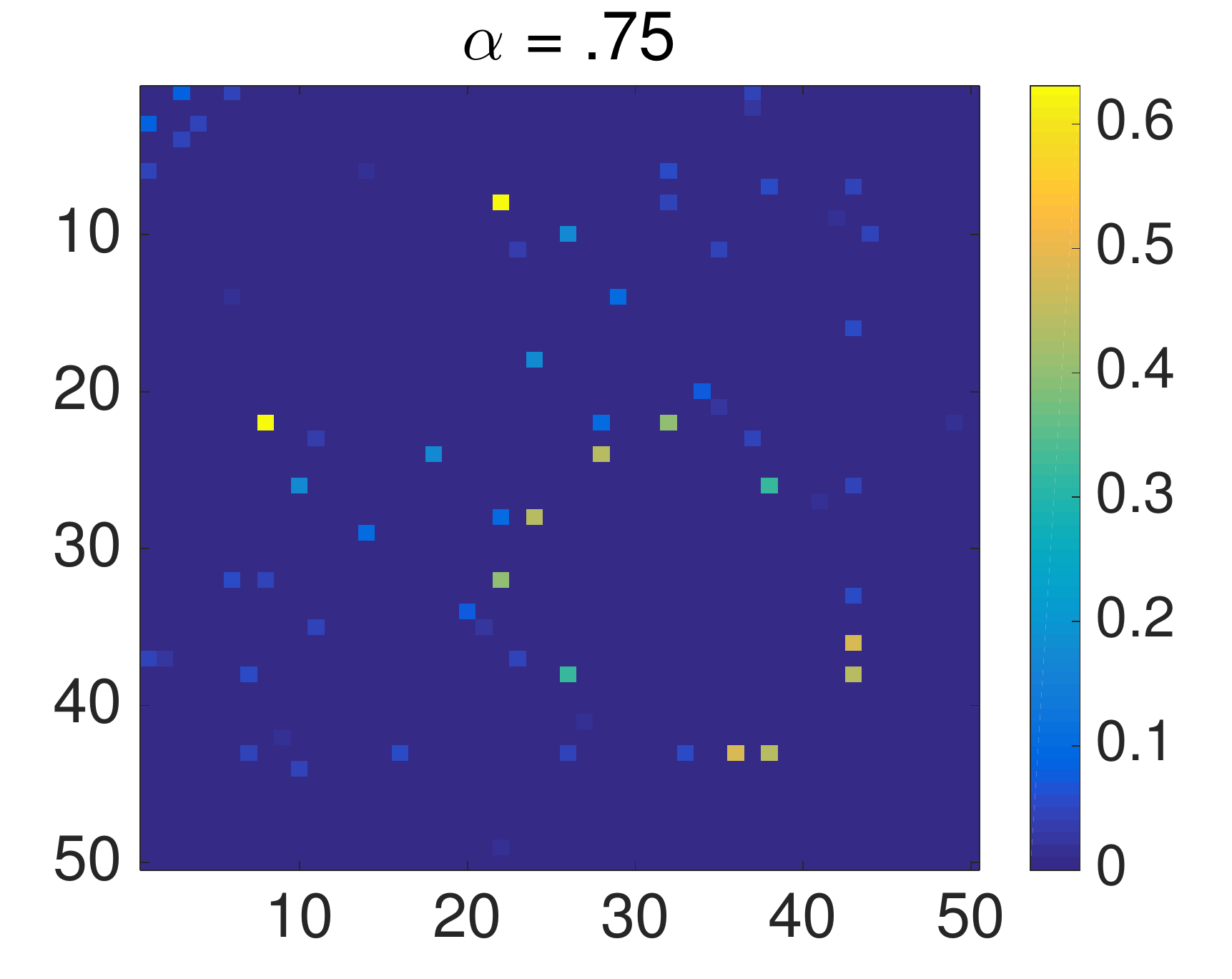}}
	\subfigure[$\hat{\boldDelta}$, KLIEP, $\alpha = 1.0$ ]{
		\label{fig.toy.kliep.100}
		\includegraphics[width=.3\textwidth]{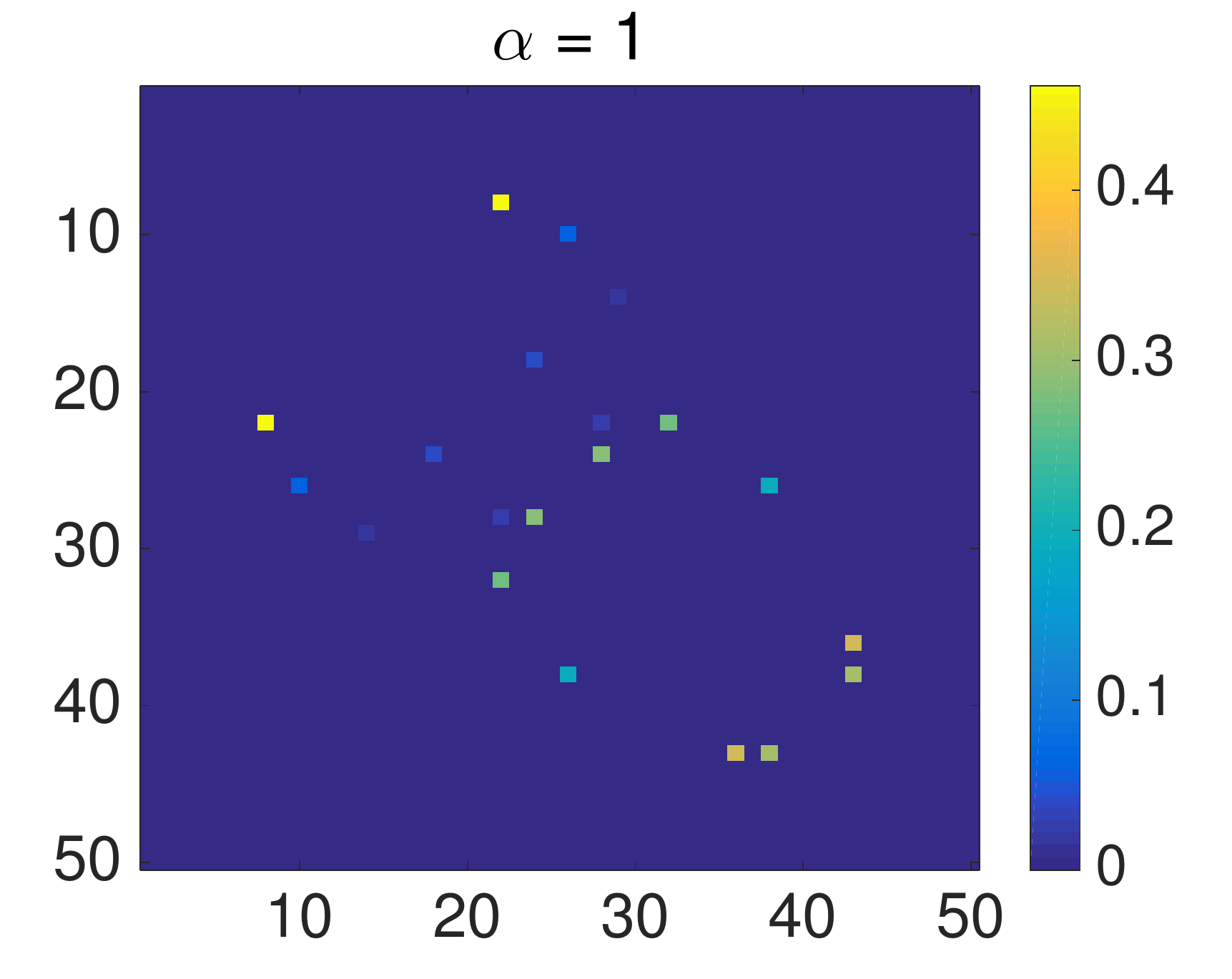}}
	\subfigure[$\hat{\boldDelta}$, KLIEP, $\alpha = 1.25$ ]{
		\label{fig.toy.kliep.125}
		\includegraphics[width=.3\textwidth]{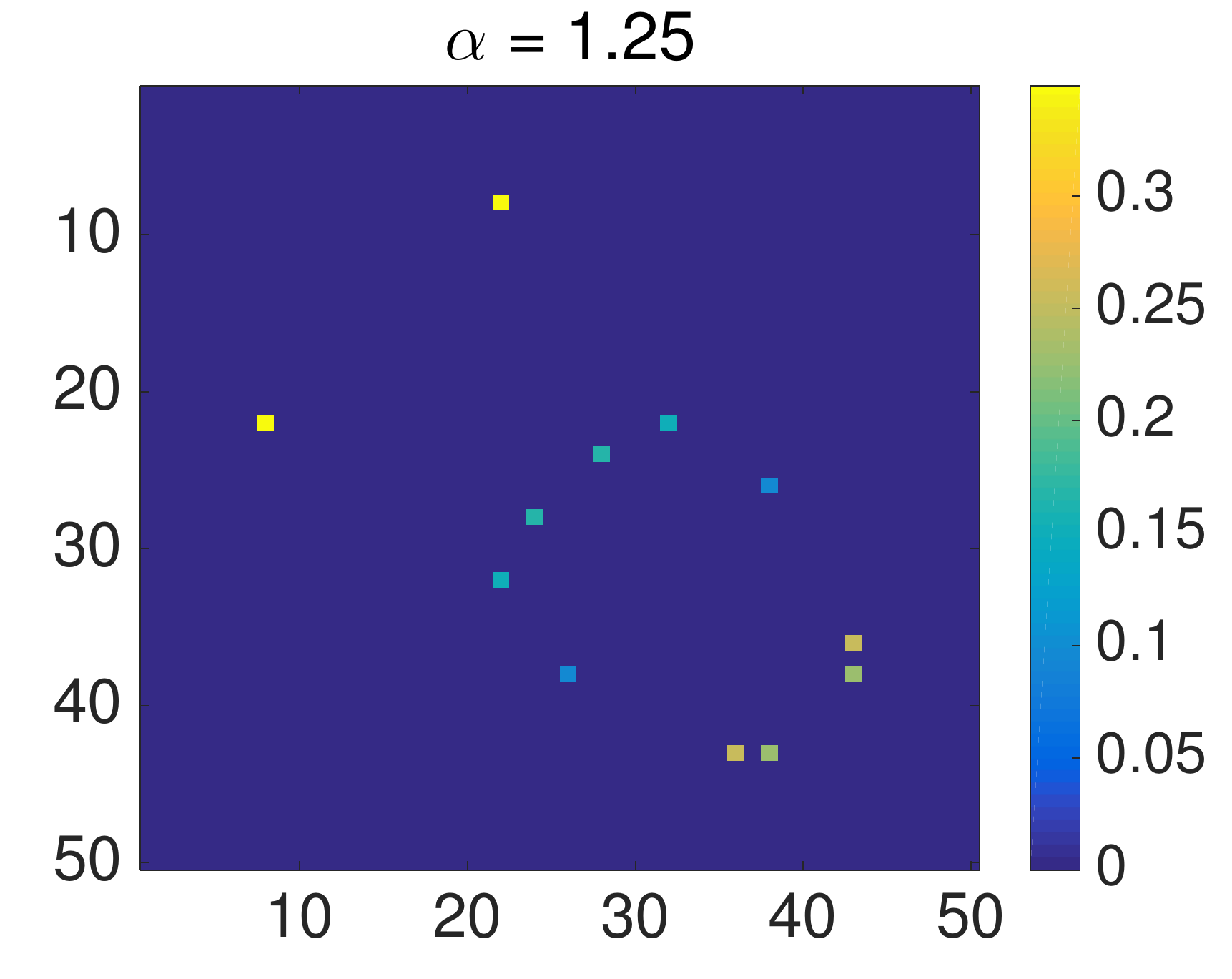}}
	\subfigure[$\hat{\boldDelta}$, CP, $\tau = 0.0$ ]{
		\label{fig.toy.cp00}
		\includegraphics[width=.3\textwidth]{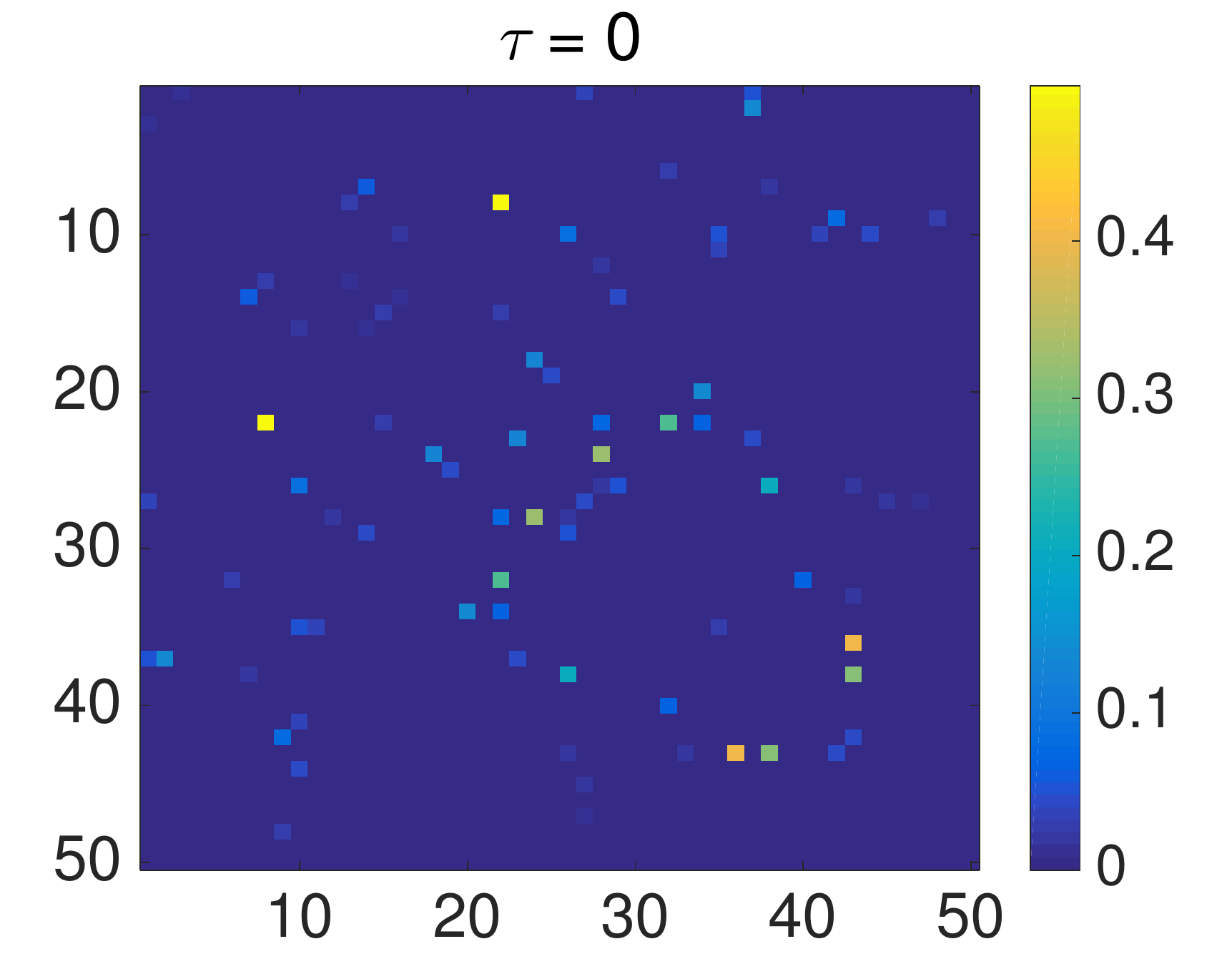}}
	\subfigure[$\hat{\boldDelta}$, CP, $\tau = 0.1$ ]{
		\label{fig.toy.cp01}
		\includegraphics[width=.3\textwidth]{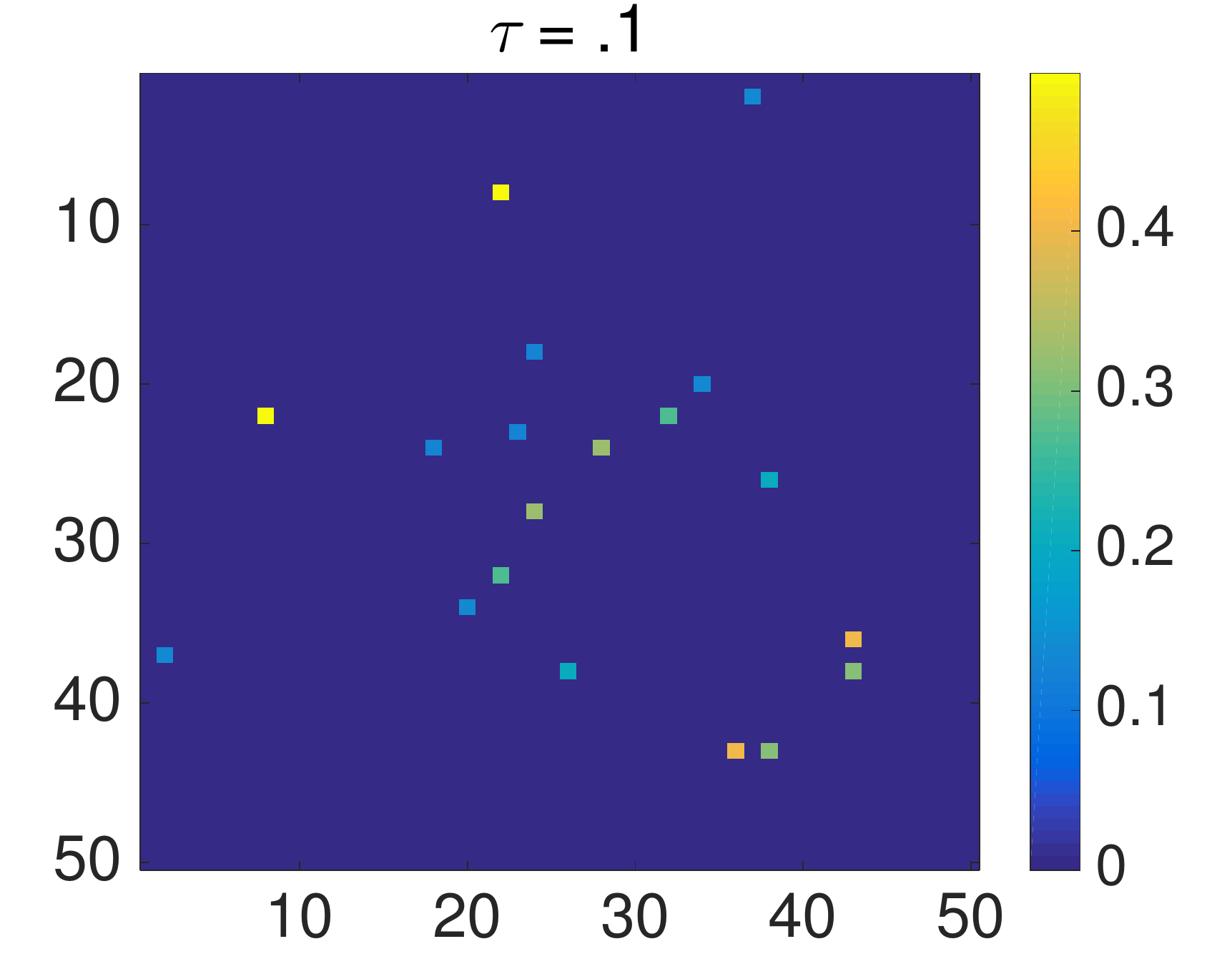}}
	\subfigure[$\hat{\boldDelta}$, CP, $\tau = 0.2$ ]{
		\label{fig.toy.cp02}
		\includegraphics[width=.3\textwidth]{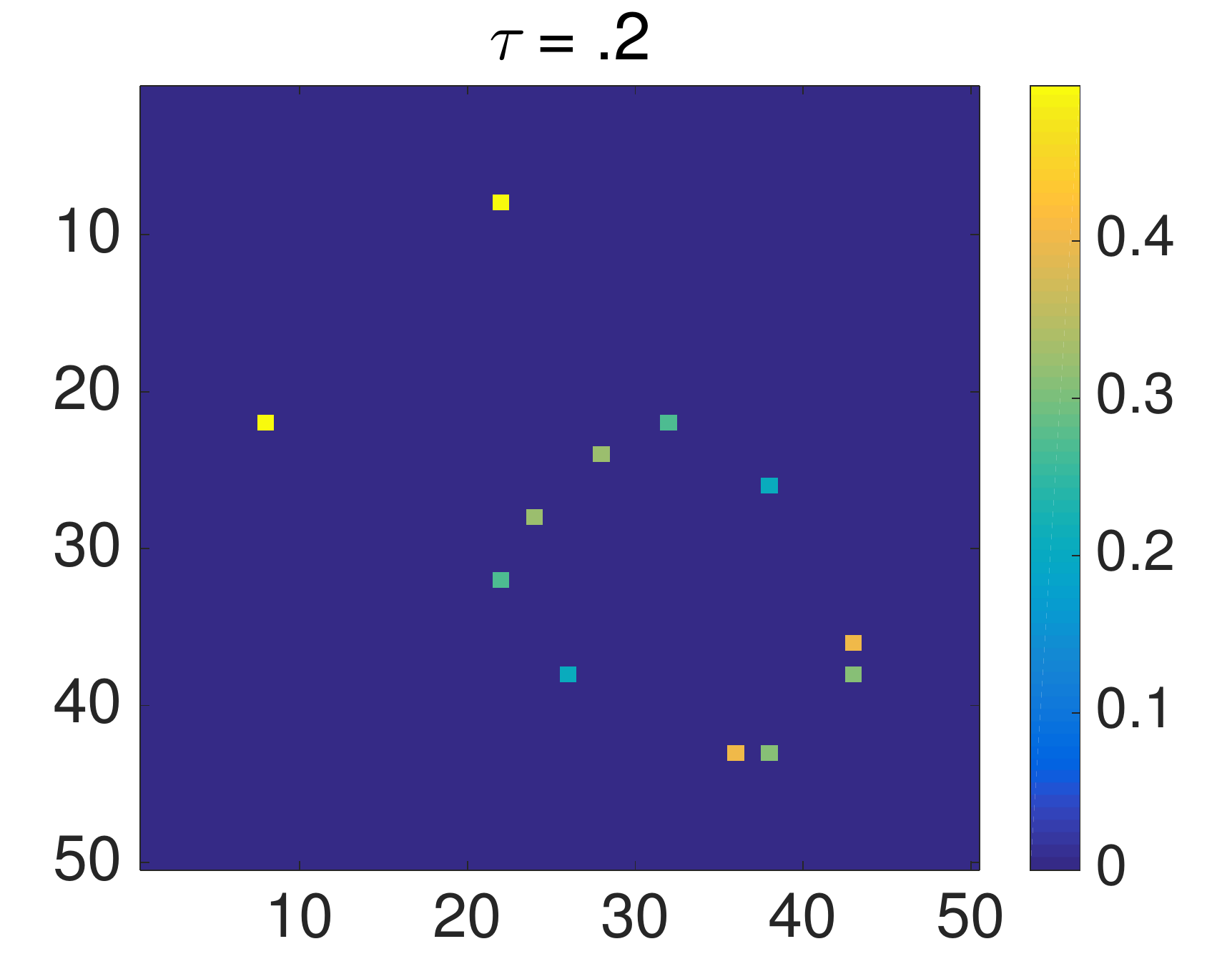}}
	\caption{Illustrative Experiments.}
\end{figure*}
\paragraph{Illustrative Example} Now we illustrate the performance of both KLIEP and CP matching using two 50 dimensional multivariate zero-mean Gaussian distributions. First, we randomly generate a $50 \times 50$ symmetric adjacency matrix $\boldA^{(P)}$ with 10\% connectivity and draw 500 samples from a Gaussian distribution with the following precision matrix:
\begin{align}
\label{eq.generate.toy}
\Theta^{(P)}_{i,j} = \begin{cases}
2 & i = j\\
0.4 & A^{(P)}_{i,j} \neq 0, i\neq j
\end{cases}
\end{align}
Then we remove 6 edges randomly from it, resulting a change pattern shown in Figure \ref{fig.toy.ground} and use it as $\boldA^{(Q)}$. Following the same step above we construct $\boldThetaQ$ and generate 500 samples again. As it was suggested by Theorem \ref{them.the.main.theorem}, we set $\lambda=\frac{\alpha \log 50}{500}$, and the learned $\hat{\boldDelta}$\footnote{We convert $\hat{\bolddelta}$ into its corresponding matrix form.} are shown in Figure \ref{fig.toy.kliep.75}, \ref{fig.toy.kliep.100} and \ref{fig.toy.kliep.125} using different $\alpha$.

The same experiments are repeated using the CP matching method. However, since the sparsity control of CP matching is via the selection of the threshold $\tau$, we set $\epsilon = 0.2$ which shows good performance empirically and plot the learned $\hat{\boldDelta}$ using different thresholds. Results are shown in Figure \ref{fig.toy.cp00}, \ref{fig.toy.cp01} and \ref{fig.toy.cp02}.

As we can see, both approaches recover the change pattern well as we increase the sparsity control parameter. 

\paragraph{ROC-curves} In this experiment, we compare two methods quantitatively using ROC curves. 
We adopt the True Positive (TP) and True Negative (TN) rate as described in \citep{Zhao2014}:
\begin{align*}
\mathrm{TPR} = \frac{\sum_{t'\in S} \delta(\hat{\bolddelta}_{
		t'} \neq \boldzero)}{\sum_{t'\in S} \delta(\bolddelta^*_{t'} \neq \boldzero)}, 	~~\mathrm{TNR} = \frac{\sum_{t'' \in S^c} \delta(\hat{\bolddelta}_{t''} = \boldzero)}{\sum_{t''\in S^c} \delta(\bolddelta^*_{t''} = \boldzero)}.
\end{align*}
We generate a adjacency matrix $A^{(P)}$ with four-neighbour lattice structure, and randomly remove $d = \sqrt{m}$ edges producing $\boldA^{(Q)}$. Two sets of $n_p = n_q = 50$ samples are generated using the same criteria mentioned in \eqref{eq.generate.toy}. The ROC curves averaged over 50 trials with different dimensions are shown in Figure \ref{fig.toy.roc}, and the AUCs are reported in Table \ref{table.auc}.

It can be seen that as the number of both dimension and changed edges increases, KLIEP method can retain stable performance while the performance of CP approach decays rapidly. 
\begin{table}
	\begin{tabular}{c|c|c|c|c|c|c|c|c}
		& $m = 9$ & $m = 16$  & $m = 25$  & $m = 36$ & $m = 49$  & $m = 64$ & $m = 81$ & $m = 100$  \\ 
		\hline 
		K & 0.8746 & 0.8865 & 0.8899 & 0.8890 & 0.8902 & 0.8878 & 0.8903 & 0.8866 \\ 
		\hline 
		CP & 0.8165 & 0.7917 & 0.7627 & 0.6829 & 0.5574 & 0.5914 & 0.5475 & 0.5656 \\ 
		\hline 
	\end{tabular} 
	\caption{The Area under the curve (AUC) of ROC plot in Figure \ref{fig.toy.roc} (``K'' for KLIEP and ``CP'' for CP matching).}
	\label{table.auc}
\end{table}
\subsection{Running Time}
Although the rigorous timing comparison is difficult due to the different implementations of KLIEP and CP matching, from our experience, KLIEP is faster but more memory-consuming as our implementation stores the entire parameter vector into the memory.
On a server with 16 Xeon cores, it takes KLIEP about 15 mins to run experiments needed for plotting Figure \ref{fig.toy.roc}, while it takes CP matching roughly 1 hour. 

As to KLIEP, we also observe that ``early stopping'' heuristics (e.g., stopping at 100 iterations) can provide an accurate non-zero pattern within a short period of time. 

\subsection{Image Difference Detection}
\begin{figure*}[t]
	\centering
	\subfigure[4:30PM, 7th Mar, 2016]{
		\label{fig.img.a2}
		\includegraphics[width=.4\textwidth]{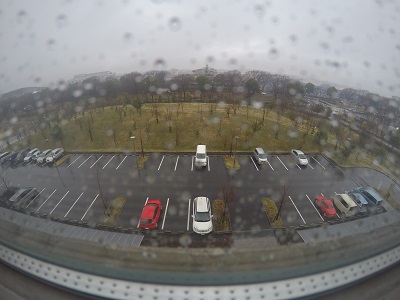}}
	\subfigure[5:30PM, 7th Mar, 2016]{
		\label{fig.park2}
		\includegraphics[width=.4\textwidth]{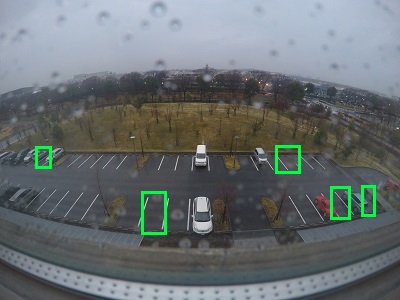}}\\
	\subfigure[Construct samples using sliding windows (red boxes). We set $\psi(\boldx_u, \boldx_v) = \exp\left(-\frac{\|\boldx_u - \boldx_v\|^2}{0.5}\right)$.]{
		\label{fig.construct.samples}
		\includegraphics[width=.4\textwidth]{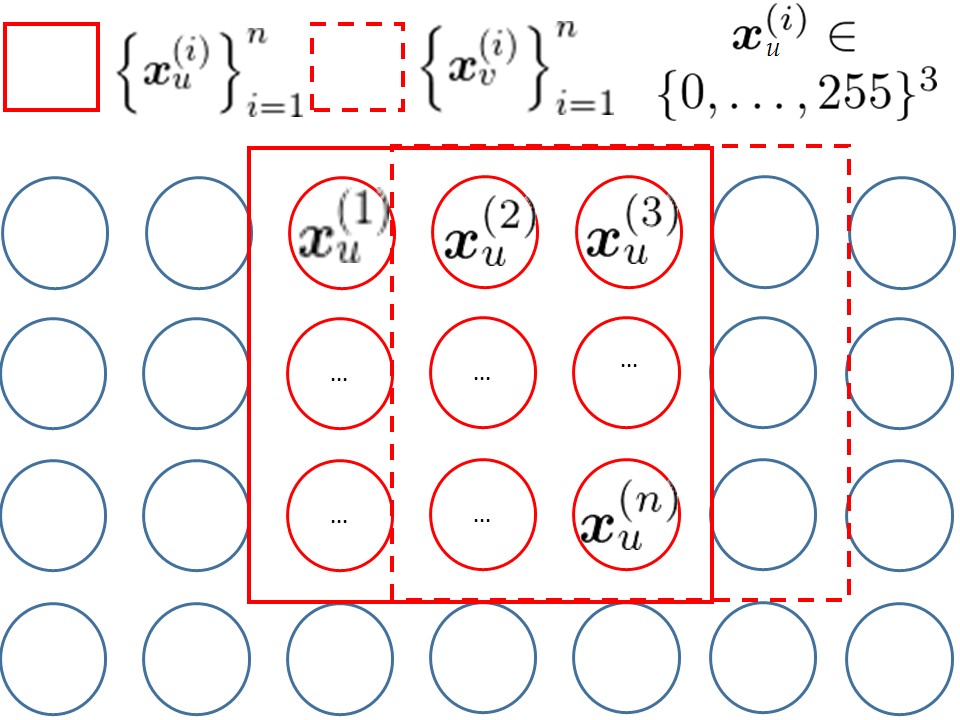}}
	\subfigure[Detected changes]{
		\label{fig.detected}
		\includegraphics[width=.4\textwidth]{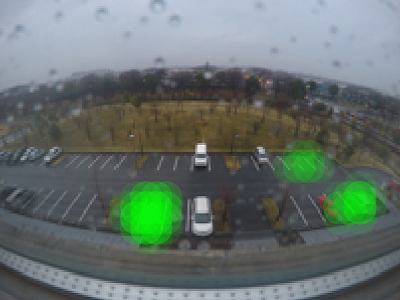}}
	\caption{Detecting changes of parking patterns from two photos.}
\end{figure*}
Two photos were taken in a rainy afternoon using a camera pointing at the parking lot of The Institute of Statistical Mathematics (ISM). In this task, we are interested in learning the changes of the parking patterns marked by green boxes in Figure \ref{fig.park2}. 
As we can see from Figure \ref{fig.img.a2} and \ref{fig.park2}, the light conditions and  positions of raindrops vary in two pictures. 

To construct samples, we use windows of pixels (Figure \ref{fig.construct.samples}). 
Each window is a dimension of a dataset, and the samples are the pixel RGB values within this window. By sliding the window across the entire picture, we may obtain samples of different dimensions. Two sets of data can be obtained by using this sample generating mechanism over two images. 

Assuming an image can be represented by an MN of windows, changes of pixels values within a window may cause changes of ``interactions'' between neighboring windows. In other words, we can discover a change by looking at the change of the dependency of pixel values between a certain window and its neighbours. 
This is more advantageous than simply looking at the pixel values  since changing the brightness of a picture may increase the pixel values in many windows simultaneously, even if the ``contrast'' between two windows does not change by much. 

By applying KLIEP on such two sets of data and highlighting adjacent window pairs that are involved in the changes of pairwise interactions, we may spot changes between two images. 
In our experiment, we use sliding windows of size $16 \times 16$ on a $200\times 150$ image, generating two sets of samples with $m=999$ and $n_p=n_q=256$. We reduce $\lambda$ until $|\hat{S}| > 40$.  The spotted changes were plotted in Figure \ref{fig.detected}. It is can be seen that KLIEP has correctly labelled almost all changed parkings between two images except one missing on the left. 

Note that here we set $\psi(\boldx_u, \boldx_v) = \exp\left(-\frac{\|\boldx_u - \boldx_v\|^2}{0.5}\right)$, and the underlying MN is highly non-Gaussian so CP matching cannot be applied here.

\section{Open Problems}
\label{sec.open}
Although pioneering works have been conducted in this area, there are still important unsolved open problems. In this section, we list a few examples. 

\paragraph{Generalized Covariance-Precision Matching} In Section \ref{sec.matching}, we introduced an equality between Gaussian covariance and precision matrix \eqref{eq.quasi.log.likelihood}. This leads to a \emph{direct} sparse change learning approach. However, it does not apply to more general pairwise MNs. A natural question is, can we extend this relationship between covariance and precision matrices to a more general principle? Particularly, in a recent work \citep{Loh2013}, the \emph{generalized covariance matrix} was used to learn a non-Gaussian graphical model structure. Would a generalized equality \eqref{eq.quasi.log.likelihood} provides us with a universal framework of learning changes between MNs? 

\paragraph{Learning Changes from Multiple MNs} In this paper, we have only reviewed the algorithms that learn changes between two MNs. In fact, in some applications, datasets may be obtained as multiple ``snapshots''. For example, gene activities may be measured at a few different time points. Under the same assumption that changes between adjacent time points are ``mild'' and ``sparse'', can we perform multiple change detections in one shot?

\paragraph{Asymmetry versus Symmetry} As we have pointed out in Section \ref{sec.summary}, there exists an asymmetry in KLIEP while Covariance-Precision matching has a symmetric formulation. An interesting future direction is to systematically investigate how such an asymmetry affects the change detection results, and more importantly, how can we automatically determine which density to be $Q$ and which one to be $P$ in the ratio formulation. 

We believe thorough investigations in these three directions will significantly expand our knowledge over the domain of learning changes between MNs in the future.

\section{Conclusion}
\label{sec.concl}
In this paper, we have reviewed an MN change learning method based on density ratio estimation and other alternative approaches. Statistical guarantees regarding the support recovery and $\ell_2$ consistency were also reported and compared. Through their direct modelling and theoretical results, we can see an interesting common pattern in all these methods: they work well regardless of the difficulty of learning individual MNs. 

These results are inspiring as they shed lights on a new family of methods that only learn the \emph{incremental} patterns. They show that if the change itself is simple enough, even with limited amount of information, we can have good learning performance. Compared to classic, \emph{static} pattern recognition, such methods are well-suited for analysing \emph{dynamic} datasets, where the ``absolute'' pattern is not the main interest, but learning the change itself is more valuable. 

These works have offered a new vision of research on learning changes between patterns. We believe these methods and theorems may have many potential applications in the years to come. 

\section{Acknowledgements}
We would like to thank Masashi Sugiyama, John Quinn and Michael Gutmann for their tremendous help during the development of the density ratio modelling idea. This work was partially supported by JSPS Grant-in-Aid for Scientific Research Activity Start-up 15H06823, MEXT kakenhi (25730013, 25120012,
26280009, 15H01678 and 15H05707), and JST-PRESTO. Authors would like to thank anonymous reviewers for their helpful comments. We would like to thank anonymous reviewers and Matthew Ames for their helpful comments and suggestions on this paper.

\bibliographystyle{plainnat}
\bibliography{main}

\begin{thebibliography}{43}
\providecommand{\natexlab}[1]{#1}
\providecommand{\url}[1]{\texttt{#1}}
\expandafter\ifx\csname urlstyle\endcsname\relax
  \providecommand{\doi}[1]{doi: #1}\else
  \providecommand{\doi}{doi: \begingroup \urlstyle{rm}\Url}\fi

\bibitem[Banerjee et~al.(2014)Banerjee, Chen, Fazayeli, and
  Sivakumar]{Banerjee2014}
A.~Banerjee, S.~Chen, F.~Fazayeli, and V.~Sivakumar.
\newblock Estimation with norm regularization.
\newblock In \emph{Advances in Neural Information Processing Systems 26}, pages
  1556--1564, 2014.

\bibitem[Banerjee et~al.(2008)Banerjee, {El Ghaoui}, and
  {d'Aspremont}]{Banerjee2008}
O.~Banerjee, L.~{El Ghaoui}, and A.~{d'Aspremont}.
\newblock Model selection through sparse maximum likelihood estimation for
  multivariate {G}aussian or binary data.
\newblock \emph{Journal of Machine Learning Research}, 9:\penalty0 485--516,
  2008.

\bibitem[Beck and Teboulle(2009)]{Beck2009}
A.~Beck and M.~Teboulle.
\newblock A fast iterative shrinkage-thresholding algorithm for linear inverse
  problems.
\newblock \emph{SIAM Journal on Imaging Sciences}, 2\penalty0 (1):\penalty0
  183--202, 2009.

\bibitem[Boyd and Vandenberghe(2004)]{Boyd2004}
S.~Boyd and L.~Vandenberghe.
\newblock \emph{Convex optimization}.
\newblock Cambridge University Press, 2004.

\bibitem[Boyd et~al.(2011)Boyd, Parikh, Chu, Peleato, and Eckstein]{Boyd2011}
S.~Boyd, N.~Parikh, E.~Chu, B.~Peleato, and J.~Eckstein.
\newblock Distributed optimization and statistical learning via the alternating
  direction method of multipliers.
\newblock \emph{Foundations and Trends{\textregistered} in Machine Learning},
  3\penalty0 (1):\penalty0 1--122, 2011.

\bibitem[Chandrasekaran et~al.(2012)Chandrasekaran, Recht, Parrilo, and
  Willsky]{Chandrasekaran2012}
V.~Chandrasekaran, B.~Recht, P.~A Parrilo, and A.~S Willsky.
\newblock The convex geometry of linear inverse problems.
\newblock \emph{Foundations of Computational mathematics}, 12\penalty0
  (6):\penalty0 805--849, 2012.

\bibitem[Chickering(1996)]{Chickering1996}
D.~M Chickering.
\newblock Learning bayesian networks is {NP}-complete.
\newblock In \emph{Learning from data}, pages 121--130. Springer, 1996.

\bibitem[Chow and Liu(1968)]{Chow1968}
C.~Chow and C.~Liu.
\newblock Approximating discrete probability distributions with dependence
  trees.
\newblock \emph{IEEE Transactions on Information Theory}, 14\penalty0
  (3):\penalty0 462--467, 1968.

\bibitem[Fazayeli and Banerjee(2016)]{Fazayeli2016}
F.~Fazayeli and A.~Banerjee.
\newblock Generalized direct change estimation in ising model structure.
\newblock 2016.

\bibitem[Friedman et~al.(2008)Friedman, Hastie, and Tibshirani]{Friedman2008}
J.~Friedman, T.~Hastie, and R.~Tibshirani.
\newblock Sparse inverse covariance estimation with the graphical {Lasso}.
\newblock \emph{Biostatistics}, 9\penalty0 (3):\penalty0 432--441, 2008.

\bibitem[Geman and Geman(1984)]{Geman1984}
S.~Geman and D.~Geman.
\newblock Stochastic relaxation, gibbs distributions, and the bayesian
  restoration of images.
\newblock \emph{IEEE Transactions on Pattern Analysis and Machine
  Intelligence}, \penalty0 (6):\penalty0 721--741, 1984.

\bibitem[Hammersley and Clifford(1971)]{Hammersley1971}
J.~M. Hammersley and P.~Clifford.
\newblock {Markov} fields on finite graphs and lattices.
\newblock Unplished, 1971.

\bibitem[Hastie et~al.(2015)Hastie, Tibshirani, and Wainwright]{Hastie2015}
Trevor Hastie, Robert Tibshirani, and Martin Wainwright.
\newblock \emph{Statistical learning with sparsity: the {Lasso} and
  generalizations}.
\newblock CRC Press, 2015.

\bibitem[Kanamori et~al.(2010)Kanamori, Suzuki, and Sugiyama]{Kanamori2010}
T.~Kanamori, T.~Suzuki, and M.~Sugiyama.
\newblock Theoretical analysis of density ratio estimation.
\newblock \emph{IEICE Transactions on Fundamentals of Electronics,
  Communications and Computer Sciences}, E93-A\penalty0 (4):\penalty0 787--798,
  2010.

\bibitem[Kolar and Xing(2012)]{Kolar2012}
M.~Kolar and E.~P. Xing.
\newblock Estimating networks with jumps.
\newblock \emph{Electronic Journal of Statistics}, 6:\penalty0 2069--2106,
  2012.

\bibitem[Kolar et~al.(2010)Kolar, Song, Ahmed, and Xing]{Kolar2010}
M.~Kolar, L.~Song, A.~Ahmed, and E.~P Xing.
\newblock Estimating time-varying networks.
\newblock \emph{Annals of Applied Statistics}, pages 94--123, 2010.

\bibitem[Koller and Friedman(2009)]{Koller2009}
D.~Koller and N.~Friedman.
\newblock \emph{Probabilistic Graphical Models: {P}rinciples and Techniques}.
\newblock MIT Press, 2009.

\bibitem[Ledoux and Talagrand(2013)]{Ledoux2013}
M.~Ledoux and M.~Talagrand.
\newblock \emph{Probability in Banach Spaces: isoperimetry and processes}.
\newblock Springer Science \& Business Media, 2013.

\bibitem[Liu et~al.(2009)Liu, Lafferty, and Wasserman]{Liu2009}
H.~Liu, J.~Lafferty, and L.~Wasserman.
\newblock The nonparanormal: Semiparametric estimation of high dimensional
  undirected graphs.
\newblock \emph{Journal of Machine Learning Research}, 10:\penalty0 2295--2328,
  2009.

\bibitem[Liu et~al.(2012)Liu, Han, Yuan, Lafferty, and Wasserman]{Liu2012}
H.~Liu, F.~Han, M.~Yuan, J.~Lafferty, and L.~Wasserman.
\newblock The nonparanormal skeptic.
\newblock In \emph{Proceedings of the 29th International Conference on Machine
  Learning (ICML2012)}, 2012.

\bibitem[Liu et~al.(2011)Liu, Xu, Gu, Gupta, Lafferty, and Wasserman]{Liu2011}
Han Liu, Min Xu, Haijie Gu, Anupam Gupta, John Lafferty, and Larry Wasserman.
\newblock Forest density estimation.
\newblock \emph{Journal of Machine Learning Research}, 12\penalty0
  (Mar):\penalty0 907--951, 2011.

\bibitem[Liu et~al.(2014)Liu, Quinn, Gutmann, Suzuki, and Sugiyama]{Liu2014}
S.~Liu, J.~A. Quinn, M.~U. Gutmann, T.~Suzuki, and M.~Sugiyama.
\newblock Direct learning of sparse changes in {Markov} networks by density
  ratio estimation.
\newblock \emph{Neural Computation}, 26\penalty0 (6):\penalty0 1169--1197,
  2014.

\bibitem[Liu et~al.(2017{\natexlab{a}})Liu, Suzuki, Relator, Sese, Sugiyama,
  and Fukumizu]{Liu2016}
S.~Liu, T.~Suzuki, R.~Relator, J.~Sese, M.~Sugiyama, and K.~Fukumizu.
\newblock Supplement to ``support consistency of direct sparse-change learning
  in {Markov} networks'', 2017{\natexlab{a}}.

\bibitem[Liu et~al.(2017{\natexlab{b}})Liu, Suzuki, Relator, Sese, Sugiyama,
  and Fukumizu]{Liu2016a}
S.~Liu, T.~Suzuki, R.~Relator, J.~Sese, M.~Sugiyama, and K.~Fukumizu.
\newblock Support consistency of direct sparse-change learning in {Markov}
  networks.
\newblock \emph{Annals of Statistics}, 2017{\natexlab{b}}.
\newblock to appear.

\bibitem[Loh and Wainwright(2013)]{Loh2013}
P-L Loh and M.~J Wainwright.
\newblock Structure estimation for discrete graphical models: Generalized
  covariance matrices and their inverses.
\newblock \emph{Annals of Statistics}, 41\penalty0 (6):\penalty0 3022--3049,
  2013.

\bibitem[Meinshausen and B{\"{u}}hlmann(2006)]{Meinshausen2006}
N.~Meinshausen and P.~B{\"{u}}hlmann.
\newblock High-dimensional graphs and variable selection with the {Lasso}.
\newblock \emph{Annals of Statistics}, 34\penalty0 (3):\penalty0 1436--1462,
  June 2006.

\bibitem[Mohan et~al.(2014)Mohan, London, Fazel, Witten, and Lee]{Mohan2014}
K.~Mohan, P.~London, M.~Fazel, D.~M Witten, and S.~Lee.
\newblock Node-based learning of multiple gaussian graphical models.
\newblock \emph{Journal of Machine Learning Research}, 15\penalty0
  (1):\penalty0 445--488, 2014.

\bibitem[Negahban et~al.(2009)Negahban, Yu, Wainwright, and
  Ravikumar]{Negahban2009}
S.~Negahban, B.~Yu, M.~J Wainwright, and P.~K Ravikumar.
\newblock A unified framework for high-dimensional analysis of $ m $-estimators
  with decomposable regularizers.
\newblock In \emph{Advances in Neural Information Processing Systems 21}, pages
  1348--1356, 2009.

\bibitem[{R Core Team}(2016)]{RCoreTeam2016}
{R Core Team}.
\newblock \emph{R: A Language and Environment for Statistical Computing}.
\newblock R Foundation for Statistical Computing, Vienna, Austria, 2016.
\newblock URL \url{https://www.R-project.org/}.

\bibitem[Ravikumar et~al.(2010)Ravikumar, Wainwright, and
  Lafferty]{Ravikumar2010}
P.~Ravikumar, M.~J. Wainwright, and J.~D. Lafferty.
\newblock High-dimensional {I}sing model selection using $\ell_1$-regularized
  logistic regression.
\newblock \emph{Annals of Statistics}, 38\penalty0 (3):\penalty0 1287--1319,
  2010.

\bibitem[Robert and Casella(2005)]{Robert2005}
C.~P. Robert and G.~Casella.
\newblock \emph{{Monte Carlo} Statistical Methods}.
\newblock Springer-Verlag, 2005.

\bibitem[Spirtes et~al.(2000)Spirtes, Glymour, and Scheines]{Spirtes2000}
P.~Spirtes, C.~N Glymour, and R.~Scheines.
\newblock \emph{Causation, prediction, and search}.
\newblock MIT press, 2000.

\bibitem[Sugiyama et~al.(2008)Sugiyama, Nakajima, Kashima, von B{\"{u}}nau, and
  Kawanabe]{Sugiyama2008a}
M.~Sugiyama, S.~Nakajima, H.~Kashima, P.~von B{\"{u}}nau, and M.~Kawanabe.
\newblock Direct importance estimation with model selection and its application
  to covariate shift adaptation.
\newblock In \emph{Advances in Neural Information Processing Systems 20}. 2008.

\bibitem[Sugiyama et~al.(2012)Sugiyama, Suzuki, and Kanamori]{Sugiyama2012}
M.~Sugiyama, T.~Suzuki, and T.~Kanamori.
\newblock \emph{Density Ratio Estimation in Machine Learning}.
\newblock Cambridge University Press, 2012.

\bibitem[Tibshirani(1996)]{Tibshirani1996}
R.~Tibshirani.
\newblock Regression shrinkage and selection via the {Lasso}.
\newblock \emph{Journal of the Royal Statistical Society. Series B
  (Methodological)}, pages 267--288, 1996.

\bibitem[Tibshirani et~al.(2005)Tibshirani, Saunders, Rosset, Zhu, and
  Knight]{Tibshirani2005}
R.~Tibshirani, M.~Saunders, S.~Rosset, J.~Zhu, and K.~Knight.
\newblock Sparsity and smoothness via the fused {Lasso}.
\newblock \emph{Journal of the Royal Statistical Society: Series B (Statistical
  Methodology)}, 67\penalty0 (1):\penalty0 91--108, 2005.

\bibitem[Tomioka and Suzuki(2014)]{Tomioka2014}
R.~Tomioka and T.~Suzuki.
\newblock Spectral norm of random tensors.
\newblock \emph{arXiv preprint arXiv:1407.1870}, 2014.

\bibitem[Tsuboi et~al.(2009)Tsuboi, Kashima, Hido, Bickel, and
  Sugiyama]{Tsuboi2009}
Y.~Tsuboi, H.~Kashima, S.~Hido, S.~Bickel, and M.~Sugiyama.
\newblock Direct density ratio estimation for large-scale covariate shift
  adaptation.
\newblock \emph{Journal of Information Processing}, 17:\penalty0 138--155,
  2009.

\bibitem[Wainwright(2009)]{Wainwright2009}
M.~J. Wainwright.
\newblock Sharp thresholds for high-dimensional and noisy sparsity recovery
  using l1-constrained quadratic programming ({Lasso}).
\newblock \emph{IEEE Transactions on Information Theory}, 55\penalty0
  (5):\penalty0 2183--2202, 2009.

\bibitem[Yuan and Lin(2006)]{Yuan2006}
M.~Yuan and Y.~Lin.
\newblock Model selection and estimation in regression with grouped variables.
\newblock \emph{Journal of the Royal Statistical Society: Series B (Statistical
  Methodology)}, 68\penalty0 (1):\penalty0 49--67, 2006.

\bibitem[Zhang and Wang(2010)]{Zhang2010}
B.~Zhang and Y.J. Wang.
\newblock Learning structural changes of {G}aussian graphical models in
  controlled experiments.
\newblock In \emph{Proceedings of the Twenty-Sixth Conference on Uncertainty in
  Artificial Intelligence (UAI2010)}, pages 701--708, 2010.

\bibitem[Zhao and Yu(2006)]{Zhao2006}
P.~Zhao and B.~Yu.
\newblock On model selection consistency of {Lasso}.
\newblock \emph{Journal of Machine Learning Research}, 7:\penalty0 2541--2563,
  2006.

\bibitem[Zhao et~al.(2014)Zhao, Cai, and Li]{Zhao2014}
S.~Zhao, T.~Cai, and H.~Li.
\newblock Direct estimation of differential networks.
\newblock \emph{Biometrika}, 101\penalty0 (2):\penalty0 253--268, 2014.

\end{thebibliography}

\end{document}